\documentclass[letterpaper]{article} 
\usepackage{paper}
\usepackage[hyphens]{url}  
\usepackage{graphicx} 
\usepackage{natbib}  
\usepackage{caption} 
\DeclareCaptionStyle{ruled}{labelfont=normalfont,labelsep=colon,strut=off} 

\usepackage{booktabs}
\usepackage{amssymb} 
\usepackage{multirow}

\newcommand{\weblink}[2]{\pdfstartlink attr{/Border [0 0 0]} user{/Subtype /Link /A << /S /URI /URI (#1) >>}\textcolor{blue}{#2}\pdfendlink}

\title{SeqLoc: Beyond the Single Frame for Cross-View Geo-Localization in Feature-Sparse Scenes}
\author{
    Junwei Zheng\textsuperscript{\rm 1},
    Yun Huang\textsuperscript{\rm 1},
    Ruize Dai\textsuperscript{\rm 1},
    Ruiping Liu\textsuperscript{\rm 1},
    Yufan Chen\textsuperscript{\rm 1},
    Kunyu Peng\textsuperscript{\rm 1},
    Kailun Yang\textsuperscript{\rm 2},
    Jiaming Zhang\textsuperscript{\rm 2,}\setcounter{footnote}{1}\thanks{Corresponding author: jiamingzhang@hnu.edu.cn},
    Guangming Wang\textsuperscript{\rm 3},
    Olaf Wysocki\textsuperscript{\rm 3},
    Rainer Stiefelhagen\textsuperscript{\rm 1}
}
\affiliations{
    \textsuperscript{\rm 1}Karlsruhe Institute of Technology, Karlsruhe, Germany\\
    \textsuperscript{\rm 2}Hunan University, Changsha, China\\
    \textsuperscript{\rm 3}University of Cambridge, Cambridge, UK
}

\begin{document}

\makeatletter
\g@addto@macro\@maketitle{%
  \noindent\includegraphics[width=\textwidth]{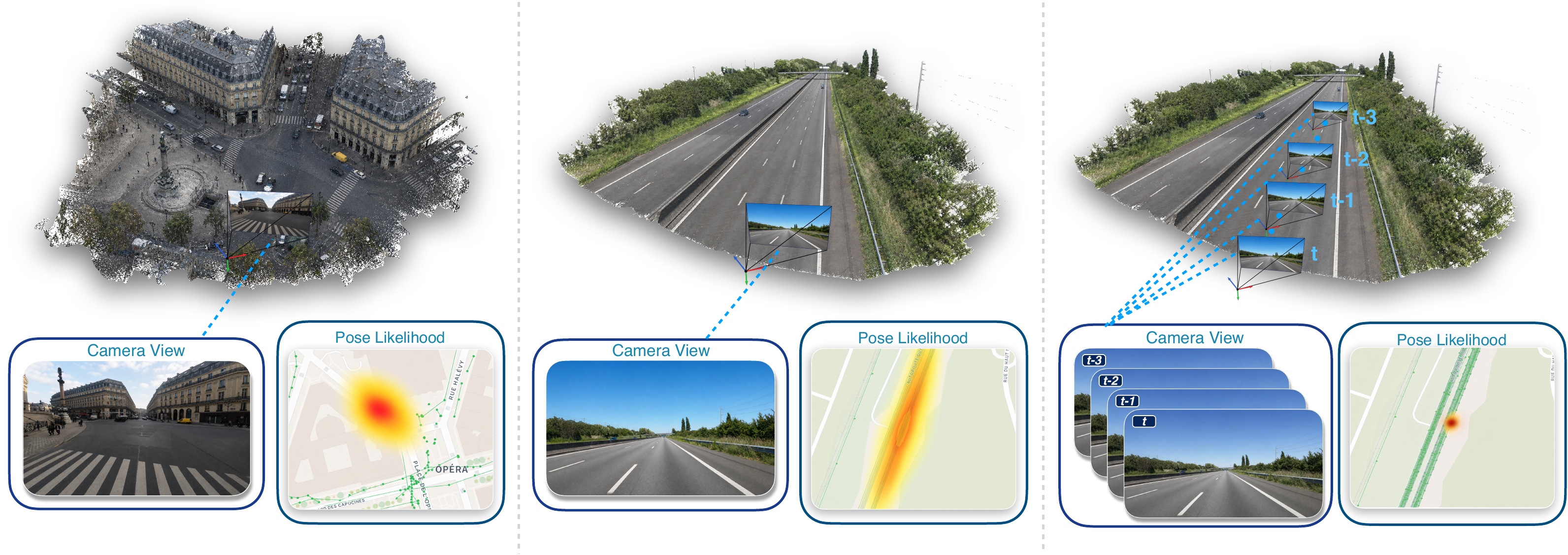}\par
  \captionof{figure}{Single-frame OSM-based Cross-View Geo-Localization is unreliable in feature-sparse scenes.
  \textbf{Left}: a feature-rich scene gives a concentrated pose likelihood.
  \textbf{Middle}: a feature-sparse scene spreads it along the road, leaving the pose ambiguous.
  \textbf{Right}: our training-free SeqLoc aggregates the likelihoods online into a sharp, accurate pose.}%
  \label{fig:teaser}%
  \vskip 1.3em%
}
\makeatother

\maketitle

\begin{abstract}
Cross-View Geo-Localization (CVGL) with OpenStreetMap (OSM) performs well in structure-rich urban environments but collapses in feature-sparse scenes such as rural roads.
To study this failure mode, in this work, we introduce CV-FSS, a benchmark that pairs sequential panoramas from five rural regions with aligned OSM maps, on which single-frame methods degrade drastically.
We then propose SeqLoc, an online test-time sequence aggregation mechanism that recursively maintains a log-belief volume with three key components:
(1) Entropy-Tempered Uncertainty (ETU) tempers each incoming pose likelihood volume by its normalized entropy;
(2) Map-Guided Relocalization (MGR) mixes a map-shaped recovery distribution into the belief so that a suppressed true pose can recover;
(3) Peak-Anchored Smoothing (PAS) derives the final pose at sub-grid precision.
Extensive experiments on CV-FSS and CV-RHO demonstrate that SeqLoc outperforms single-frame localization by a large margin, improving both position and orientation recall by \textbf{over 50\%}.
The benchmark and source code are publicly available at \weblink{https://zhengjunwei.com/publications/SeqLoc/SeqLoc.html}{SeqLoc}.
\end{abstract}

\section{Introduction}

Reliable self-localization is a prerequisite for autonomous vehicles~\cite{chalvatzaras2022survey} and mobile robots~\cite{ullah2024mobile,zheng2025scene}, yet GNSS is frequently degraded by multipath effects, occlusion, or jamming.
Cross-View Geo-Localization (CVGL) offers an appealing alternative that estimates the camera pose by matching ground-level images against geo-tagged overhead references.
Among possible references, OpenStreetMap (OSM) is particularly attractive since it is lightweight, freely available worldwide, and frequently updated by an active community.
Building on OSM, recent methods such as OrienterNet~\cite{sarlin2023orienternet} and RHO~\cite{zheng2026rho} reach meter-level accuracy by matching neural bird's-eye-view (BEV) features~\cite{wei2024onebev,wei2026onebev++} against rasterized map tiles.

These successes, however, are largely confined to structure-rich urban environments.
When a vehicle drives through feature-sparse scenes, such as rural roads, forest corridors, and open farmland, single-frame localization becomes unreliable.
Our analysis shows that the dominant failure is not inaccurate depth or scale but the ambiguity of the observation itself.
Along a straight road with few distinctive structures, the pose likelihood forms a long ridge in the driving direction, and the resulting wrong peaks can lie far away from the true pose.
Yet no existing benchmark isolates and measures this failure mode.

\begin{figure}[t]
\centering
\includegraphics[width=\columnwidth]{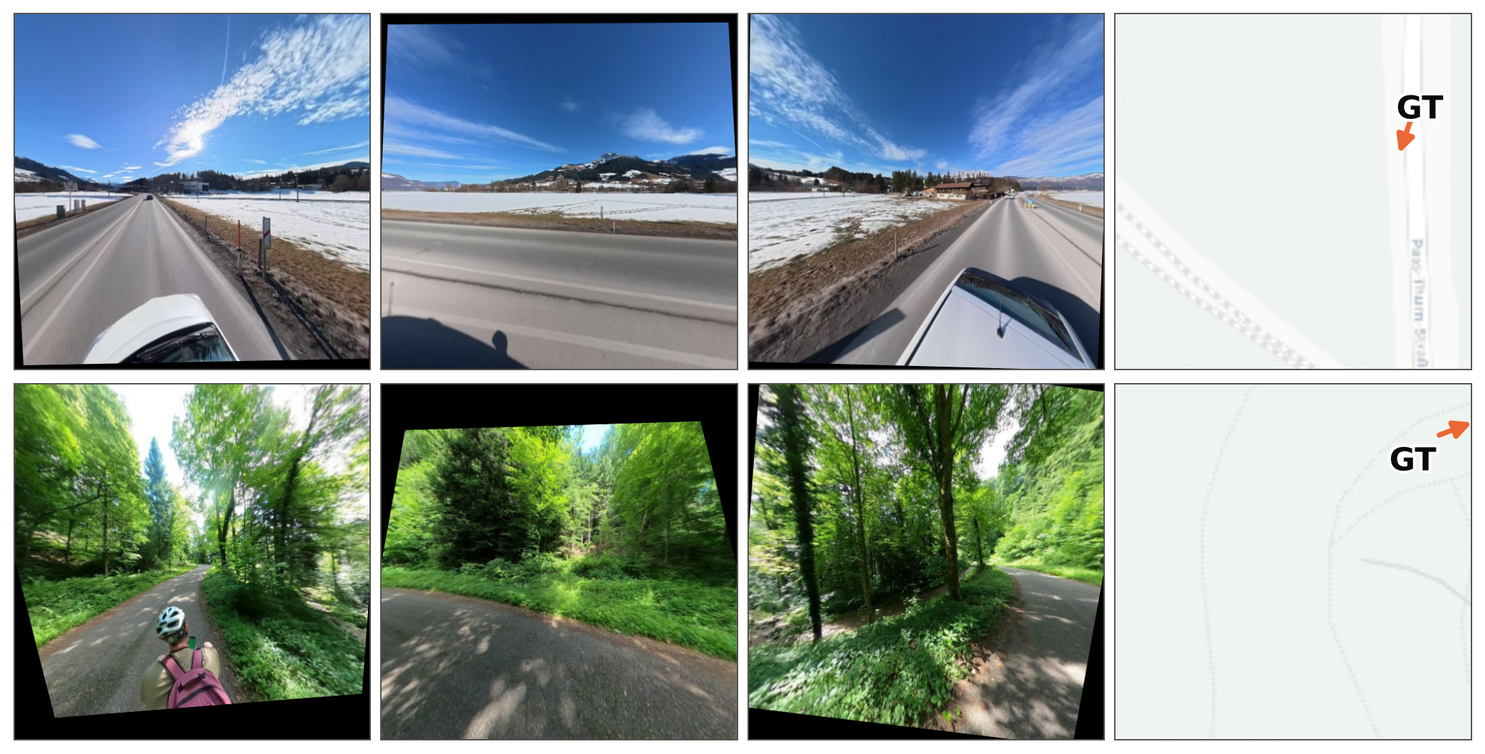}
\caption{Data samples from CV-FSS.
The three views are split from one panorama and rectified using the known gravity direction.
On the paired local OSM tile, the orange arrow marks the ground-truth pose.}
\label{fig:samples}
\end{figure}

\begin{figure}[t]
\centering
\includegraphics[width=\columnwidth]{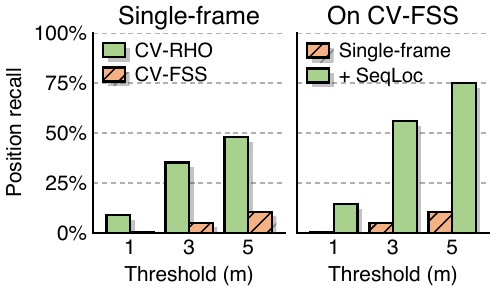}
\caption{Position recall (\%) at $1/3/5$\,m of RHO and its SeqLoc-aggregated version.
\textbf{Left}: single-frame localization degrades drastically from the urban CV-RHO to the proposed feature-sparse CV-FSS.
\textbf{Right}: on CV-FSS, SeqLoc recovers accurate poses where single-frame estimation fails.}
\label{fig:motivation}
\end{figure}

To fill this gap, we construct Cross-View Feature-Sparse Scenes (CV-FSS), a benchmark that pairs sequential panoramas from five rural regions with aligned OSM tiles. Figure~\ref{fig:samples} shows some representative samples.
Most of its routes pass through highly ambiguous road stretches, where strong localization cues appear only at sparse intersections and building clusters.
On CV-FSS, single-frame methods that perform well on the existing CV-RHO benchmark~\cite{zheng2026rho} degrade drastically, as shown on the left of Figure~\ref{fig:motivation}.

These failures motivate us to look beyond the single frame and exploit sequential observations.
Even if a single frame is ambiguous, the likelihood accumulated along the route eventually disambiguates the pose, as shown on the right of Figure~\ref{fig:motivation}.
However, training-based sequence models~\cite{zhang2023cross} require dedicated architectures and sequence annotations, while inference-time fusion in current metric methods~\cite{sarlin2023orienternet,zheng2026rho} operates offline and requires access to future frames.
Worse, naive recursive fusion is structurally fragile.
Once a few confident but wrong likelihoods suppress the true pose, multiplicative updates can hardly recover it~\cite{thrun2001robust}, and a single locked trajectory can ruin the overall performance.
In addition, consecutive frames in this setting are separated by large distances~\cite{anguelov2010google}, so feature-level cross-frame matching is unreliable, and aggregation should instead operate on the pose likelihood itself.

We therefore propose SeqLoc, an online test-time sequence aggregation mechanism that turns a stream of pose likelihood volumes into accurate sequential estimates, as illustrated in Figure~\ref{fig:teaser}.
SeqLoc maintains a log-belief volume recursively with three key components:
(1) Entropy-Tempered Uncertainty (ETU) tempers each incoming pose likelihood volume by its normalized entropy, which down-weights ambiguous frames relative to discriminative ones, as shown in Figure~\ref{fig:etu};
(2) Map-Guided Relocalization (MGR) mixes a small map-shaped recovery distribution into the belief at every step, providing a probabilistic escape route that lets a suppressed true pose recover within a few frames instead of locking the whole trajectory permanently;
(3) Peak-Anchored Smoothing (PAS) derives the final pose at sub-grid precision through a local expectation anchored at the belief peak.
All three components operate purely on probability volumes at inference time.
They are training-free, backbone-agnostic, and add negligible overhead.

In summary, our contributions are listed as follows:
\begin{itemize}
\item We introduce CV-FSS, a benchmark dedicated to OSM-based metric CVGL in feature-sparse scenes, covering five rural regions with sequential panoramas aligned to OSM tiles.
\item We propose SeqLoc, an online test-time sequence aggregation mechanism with three training-free and backbone-agnostic components, namely Entropy-Tempered Uncertainty, Map-Guided Relocalization, and Peak-Anchored Smoothing.
\item Extensive experiments on CV-FSS and CV-RHO demonstrate that SeqLoc outperforms single-frame CVGL by a large margin, improving both position and orientation recall by \textbf{over 50\%}.
\end{itemize}

\begin{figure*}[t]
\centering
\includegraphics[width=\textwidth]{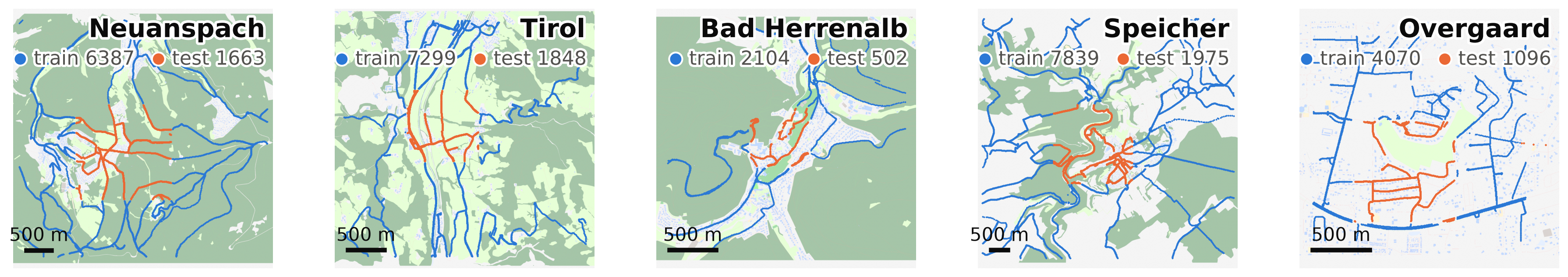}
\caption{Overview of the five regions in CV-FSS.
On the OSM of each region, training and test routes are marked in blue and orange, and the legends list the corresponding numbers of panoramic frames.}
\label{fig:dataset}
\end{figure*}

\section{Related Work}
\subsection{Cross-View Geo-Localization}
CVGL estimates the geographic location of a ground-level image by matching it against geo-tagged overhead references.
Early efforts formulate the task as large-area retrieval, where the query is matched against a database of satellite patches covering the region of interest~\cite{hu2018cvm,liu2019lending,shi2019spatial,yang2021cross,zhu2022transgeo,deuser2023sample4geo,fervers2024statewide,ye2025where}.
Since retrieval only yields patch-level positions, later works pursue metric localization, which recovers the fine-grained camera pose within a reference patch~\cite{zhu2021vigor,xia2022visual,shi2022beyond,lentsch2023slicematch,wang2023fine,fervers2023uncertainty,xia2023convolutional,xia2025fg}.
In terms of reference data, the above methods rely on high-resolution satellite imagery, which is costly to acquire and update.
OpenStreetMap, a freely available and globally maintained vector map, has thus emerged as an attractive alternative.
OrienterNet~\cite{sarlin2023orienternet} matches neural BEV features against rasterized OSM tiles, MapLocNet~\cite{wu2024maplocnet} registers visual and map features in a coarse-to-fine manner,
OSMLoc~\cite{liao2026osmloc} boosts the localization accuracy using fused geometric and semantic guidance, and
RHO~\cite{zheng2026rho} exploits holistic panoramic imagery~\cite{zheng2024open,hu2024deformable,fan2026more} for robust OSM-based localization under adverse conditions.
It's worth noting that all aforementioned methods focus on feature-rich scenes.
CVGL in feature-sparse scenes is still underexplored.
To fill this gap, we construct a benchmark termed CV-FSS, pairing sequential panoramas from five rural regions with aligned OSM tiles to study OSM-based metric CVGL in feature-sparse scenes.

\subsection{Sequence-Based Localization}
Sequential observations provide complementary visual cues that a single frame cannot offer, and existing work exploits them at two different stages.
The first line integrates sequences during training, where sequence-specific architectures match a query video or image sequence against the reference~\cite{vyas2022gama,shi2022cvlnet,zhang2023cross,regmi2021video,pillai2024garet,wu2024cross,gao2026trajectory}.
The second line aggregates visual cues at inference time.
Classical robot localization fuses a stream of observations with recursive Bayesian filtering~\cite{dellaert1999monte,thrun2002probabilistic}, visual SLAM jointly estimates the camera trajectory and a map from the same stream~\cite{cadena2016past,mur2016orb,cao2023tightly}, and SeqSLAM~\cite{milford2012seqslam} matches route segments instead of individual images to overcome appearance change.
Recent metric methods revisit this idea, yet only as an offline post-processing step that fuses per-frame likelihoods~\cite{sarlin2023orienternet,liao2026osmloc,zheng2026rho}.
Training-time approaches require sequence-specific architectures, while existing inference-time fusion in OSM-based metric CVGL relies on future frames.
In contrast, our test-time sequence aggregation runs online, relying only on past frames without access to future ones.
Moreover, it introduces three key components to achieve robust and precise localization in feature-sparse scenes.

\begin{table}[t]
\centering
\setlength{\tabcolsep}{2.5pt}
\begin{tabular}{@{}lcccccc@{}}
\toprule[1.5pt]
Dataset & OSM & Metric & Seq. & Sparse & \#Pano & \#Pin \\
\midrule
VIGOR & $\times$ & \checkmark & $\times$ & $\times$ & 105.2k & $\times$ \\
KITTI-CVL & $\times$ & \checkmark & \checkmark & $\times$ & $\times$ & 31.0k \\
MGL & \checkmark & \checkmark & \checkmark & $\times$ & $\times$ & 760.0k \\
CV-Cities & $\times$ & $\times$ & $\times$ & $\times$ & 223.7k & $\times$ \\
CV-RHO & \checkmark & \checkmark & \checkmark & $\times$ & 114.0k & 342.1k \\
CV-FSS (ours) & \checkmark & \checkmark & \checkmark & \checkmark & 34.8k & 104.4k \\
\bottomrule[1.5pt]
\end{tabular}
\caption{Comparison between the proposed CV-FSS benchmark and widely used CVGL datasets.
Seq. and Sparse denote sequential frames and feature-sparse scenes, and \#Pano and \#Pin the numbers of panoramas and pinhole images, respectively.
Among these datasets, only CV-FSS covers feature-sparse scenes.}
\label{tab:dataset}
\end{table}

\section{Methodology}
\subsection{CV-FSS Dataset}
Compared to satellite imagery, OSM is easier to acquire, updated more frequently, and requires less storage.
CV-FSS targets OSM-based metric CVGL in feature-sparse scenes.
It provides $34.8$k panoramic frames recorded as continuous routes across five regions, covering \emph{rural roads}, \emph{open fields}, and \emph{forest corridors} where buildings and other distinctive landmarks are scarce.
Every panorama is paired with a local OSM tile and split into three perspective views that are rectified with the known gravity direction, as shown in Figure~\ref{fig:samples}.
The panoramas are collected from Mapillary, a platform of crowd-sourced street-level imagery released under the CC BY-SA 4.0 license.
The ground-truth poses come from the platform's fusion of structure-from-motion and GPS.

Within each region, we divide the routes into a training set and a test set, shown in blue and orange in Figure~\ref{fig:dataset}.
Table~\ref{tab:dataset} compares CV-FSS with VIGOR~\cite{zhu2021vigor}, KITTI-CVL~\cite{shi2022cvlnet}, MGL~\cite{sarlin2023orienternet}, CV-Cities~\cite{huang2024cv}, and CV-RHO~\cite{zheng2026rho}.
Unlike existing benchmarks that focus on feature-rich urban scenes, CV-FSS deliberately targets the underexplored feature-sparse regime.
It therefore contains fewer frames than city-scale datasets, yet single-frame localization on it is far more challenging, since each frame on its own rarely determines the pose.

\begin{figure}[t]
\centering
\includegraphics[width=\columnwidth]{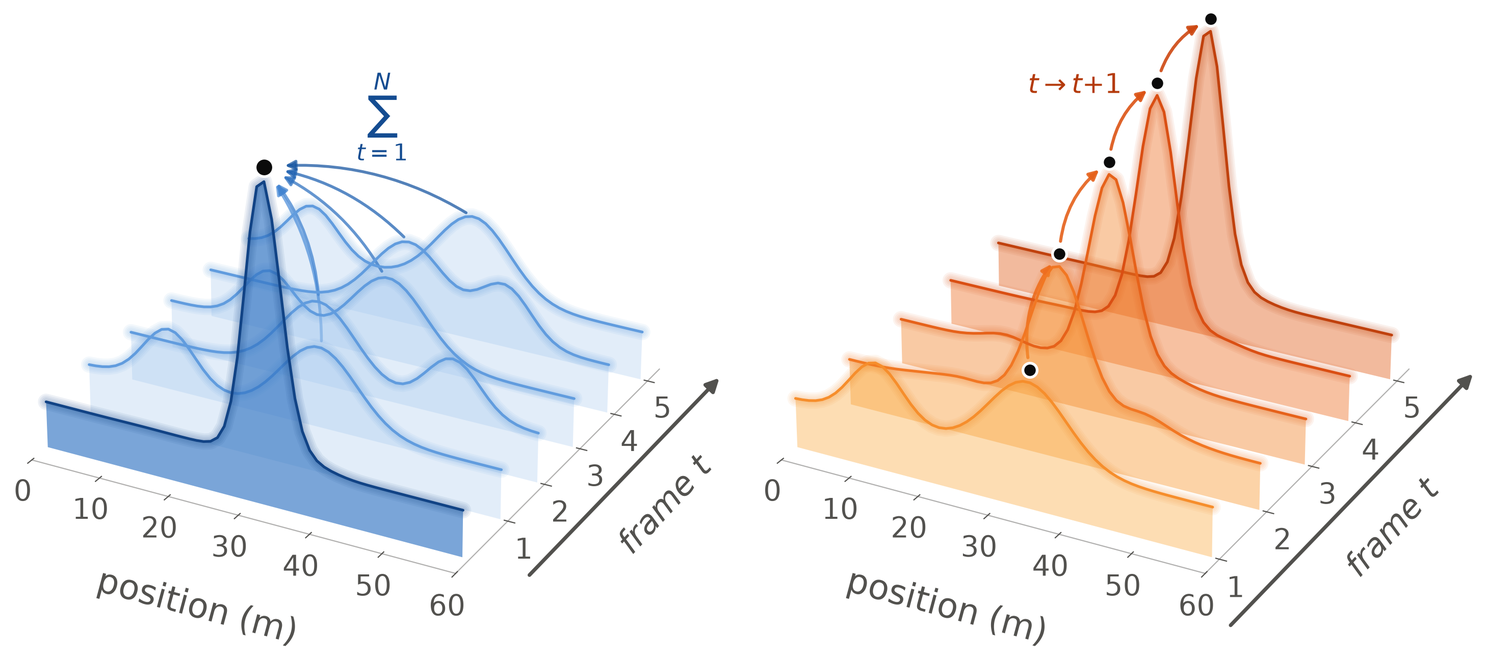}
\caption{Offline vs. online test-time sequence aggregation.
Each curve is the single-frame pose likelihood at one time step.
\textbf{Left}: offline methods sum all $N$ frames at once using future frames, localizing only the reference frame.
\textbf{Right}: SeqLoc updates the belief recursively from one frame to the next using only past frames, localizing every frame.}
\label{fig:ttsa}
\end{figure}

\subsection{Test-Time Sequence Aggregation}
\subsubsection{Superiority of SeqLoc}
Different from existing sequence-based methods, SeqLoc is a \textbf{test-time} fusion method aggregating pose likelihood volumes \textbf{without retraining}.
Other test-time aggregation methods, like OrienterNet and RHO, align all per-frame volumes to a common reference and sum them in one batch, as shown on the left of Figure~\ref{fig:ttsa}.
This relies on future frames, so it cannot run in real time and localizes only the reference frame.
SeqLoc instead aggregates the volumes \textbf{online}, as shown on the right of Figure~\ref{fig:ttsa}.
It propagates a running belief through the relative motion and fuses each incoming volume, so it \textbf{localizes every frame} as it arrives.
Metric CVGL is meant to guide a moving platform, which needs its pose at the current frame rather than one pose available only after the route ends.
Note that SeqLoc is also \textbf{backbone-agnostic}, meaning that it can be applied to any OSM-based metric CVGL backbone as long as the CVGL model outputs a pose likelihood volume.
We discretize this pose space into $M$ cells.
For frame $t$ of a sequence, the backbone outputs a pose likelihood volume $p_t(\mathbf{x},\theta)$ over this grid, which we normalize to sum to one.
SeqLoc maintains a log-belief volume $b_t(\mathbf{x},\theta)$ over the same grid, so the recursive update adds per-frame likelihoods instead of multiplying them and avoids underflow.
At every step we first propagate the previous belief to the current frame.
Let $\Delta_t$ be the relative motion read by odometry from frame $t-1$ to frame $t$ and let $T_{\Delta_t}$ be the rigid transform it induces on the pose space, so that the predicted belief is
\begin{equation}
\hat{b}_t(\mathbf{x},\theta) = b_{t-1}\big(T_{\Delta_t}^{-1}(\mathbf{x},\theta)\big) .
\label{eq:predict}
\end{equation}
Each current cell thus reads the previous belief at its source pose $T_{\Delta_t}^{-1}(\mathbf{x},\theta)$, as in backward warping.
We then fuse the new likelihood volume into the predicted belief,
\begin{equation}
b_t = \mathrm{log\,softmax}\big(\log p_t + \hat{b}_t\big) ,
\label{eq:update}
\end{equation}
where $\mathrm{log\,softmax}$ normalizes over the whole volume.
Besides the superiority of the online test-time sequence aggregation mechanism, SeqLoc has three effective components at three different stages of this recursion, namely the incoming likelihood volume (ETU), the belief after fusion (MGR), and the pose extracted from the belief (PAS).
These components push the localization accuracy to another level.
We explain their details below.

\begin{figure}[t]
\centering
\includegraphics[width=\columnwidth]{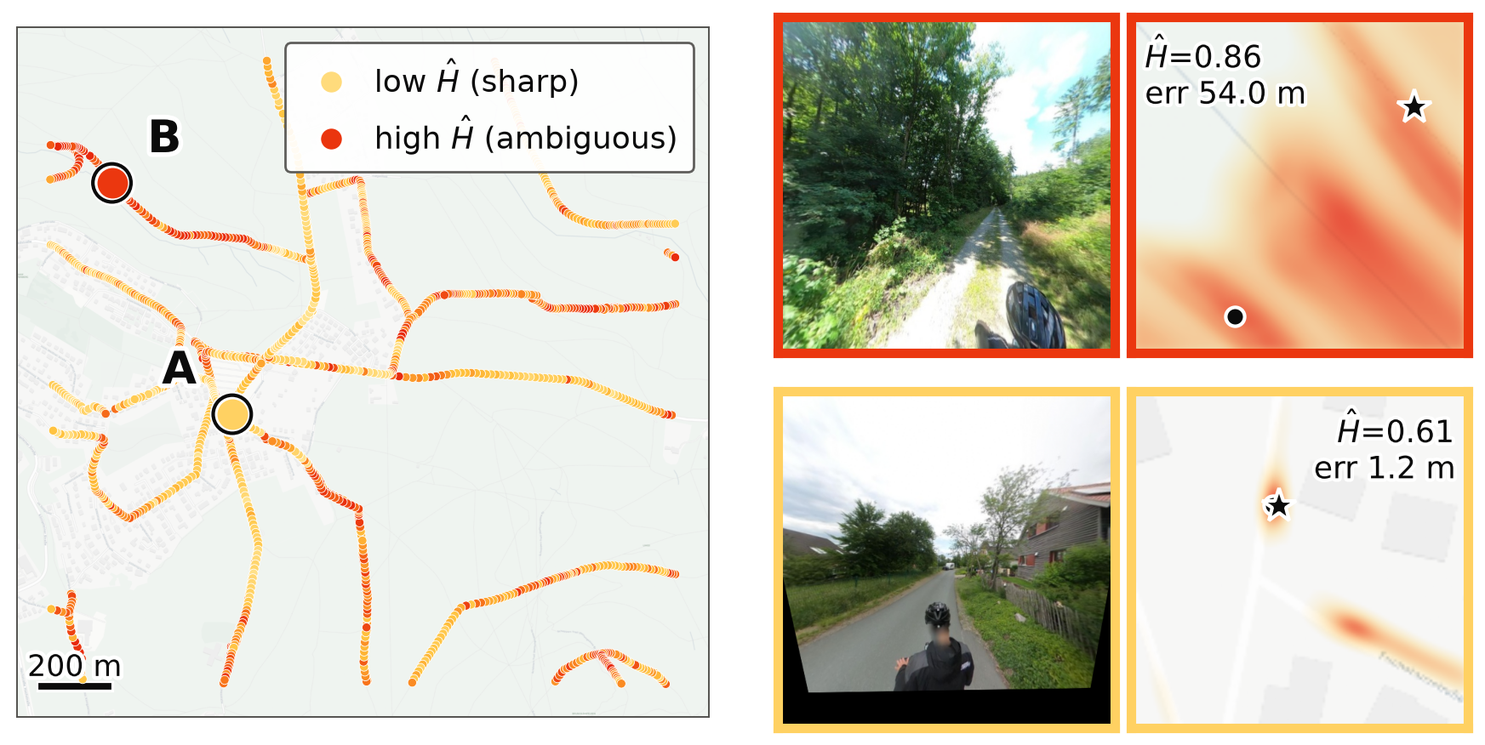}
\caption{Entropy reflects reliability.
\textbf{Left}: frames from the Neuanspach of CV-FSS, colored by the normalized entropy $\hat{H}$.
\textbf{Right}: the distinctive scene A in yellow has smaller error than the forest corridor B in red.
Pose likelihood volumes are marginalized over the orientation dimension.
Black dots indicate their argmax, while stars are the ground truth.}
\label{fig:etu}
\end{figure}

\subsubsection{Entropy-Tempered Uncertainty (ETU)}
The reliability of a single frame varies considerably along a route.
A frame that observes distinctive structure produces a peaked likelihood volume, whereas a frame recorded in a forest corridor produces a diffuse one, as shown in Figure~\ref{fig:etu}.
Crucially, a diffuse likelihood volume is \textbf{not uniform}, since it retains a ridge along the road or a spurious peak far from the true pose.
Unweighted recursive fusion gives it the same influence as a peaked one, so a small number of ambiguous frames can flatten the belief or bias it away from the true pose.
ETU instead measures how informative each likelihood volume is and tempers it accordingly before fusion.
We quantify the ambiguity of the single-frame likelihood volume $p_t$ by its entropy, normalized by the entropy of the uniform distribution over its $M$ cells,
\begin{equation}
\hat{H}_t = \frac{H(p_t)}{\log M} = \frac{-\sum_{\mathbf{x},\theta} p_t(\mathbf{x},\theta)\log p_t(\mathbf{x},\theta)}{\log M} .
\end{equation}
The normalized entropy lies in $[0,1]$ and approaches $0$ when $p_t$ is peaked and $1$ when it is nearly uniform.
We evaluate it over the full $(\mathbf{x},\theta)$ volume rather than over a spatial marginal, so that ambiguity in orientation is also considered.
We convert $\hat{H}_t$ into a per-frame weight that stays strictly positive,
\begin{equation}
w_t = \max\big(1-\hat{H}_t,\, 10^{-3}\big) .
\end{equation}
Since the belief is maintained in the log domain, we temper the likelihood volume with $w_t$ and denote the result by $\tilde{p}_t$,
\begin{equation}
\log \tilde{p}_t = w_t \log p_t = \log\big(p_t^{\,w_t}\big) .
\end{equation}
Since $w_t$ enters as an exponent on $p_t$, it acts as an inverse temperature, so a likelihood with high entropy receives a small $w_t$ at a high temperature.
The high temperature then drives the likelihood \textbf{toward uniform} and suppresses the ridge or spurious peak, so the update leaves the belief almost unchanged instead of flattening or biasing it.

\begin{figure}[t]
\centering
\includegraphics[width=\columnwidth]{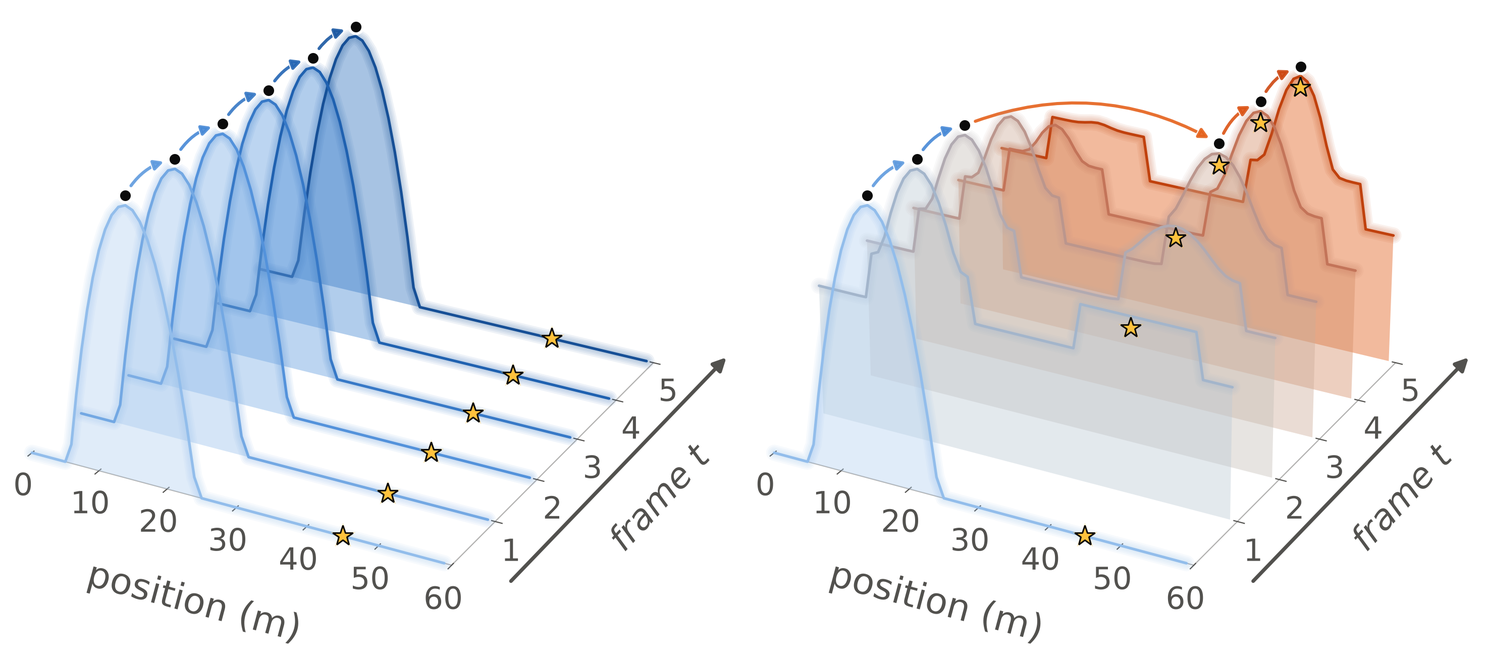}
\caption{MGR recovers a suppressed true pose.
Curves are the per-frame belief, black dots their argmax, and gold stars the ground truth.
\textbf{Left}: without MGR, the argmax locks onto a wrong peak far from the ground truth.
\textbf{Right}: with MGR, it recovers to the ground truth within a few frames.}
\label{fig:mgr}
\end{figure}

\subsubsection{Map-Guided Relocalization (MGR)}
Recursive fusion is multiplicative and therefore fragile.
Once a few wrong frames zero the belief at the true pose, later correct likelihoods multiply against that zero and cannot revive it, so the estimate locks onto a wrong peak, as shown on the left of Figure~\ref{fig:mgr}.
MGR keeps an escape route open by mixing a small recovery distribution into the belief at each step, shaped by the map rather than a uniform one.
We read each cell's distance $d(\mathbf{x})$ to the nearest road on OSM, pass it through a Gaussian of width $\sigma$, normalize over the pose volume, and add a uniform floor,
\begin{equation}
q(\mathbf{x}) = 0.9\,\exp\!\big(-d(\mathbf{x})^2/(2\sigma^2)\big)/Z + 0.1\,\mathcal{U} ,
\label{eq:road}
\end{equation}
where $Z$ normalizes the road-shaped term and $\mathcal{U}$ is uniform over the pose volume.
We then mix $q$ into the belief with a small weight $\varepsilon$,
\begin{equation}
b_t \leftarrow \log\!\big((1-\varepsilon)\exp b_t + \varepsilon\, q\big) .
\label{eq:mgr}
\end{equation}
Since this floor rises along the road, where a pose most likely appears, a suppressed true pose is able to recover within a few frames, as shown on the right of Figure~\ref{fig:mgr}.

\begin{figure}[t]
\centering
\includegraphics[width=\columnwidth]{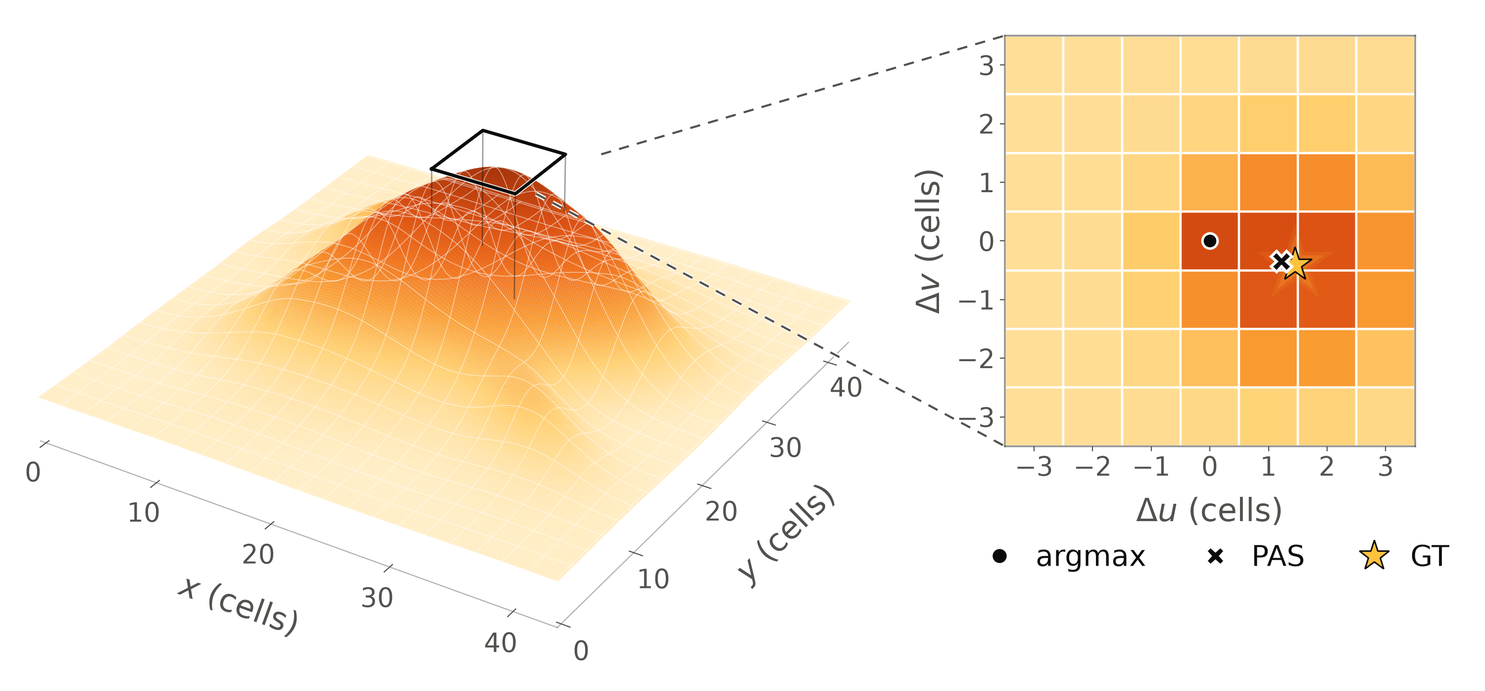}
\caption{PAS refines the pose to sub-grid precision.
\textbf{Left}: the belief marginalized over orientation, with the black box marking the local window around its peak.
\textbf{Right}: inside that window, the discrete argmax stays on a grid cell while the PAS expectation shifts toward the ground truth.}
\label{fig:pas}
\end{figure}

\subsubsection{Peak-Anchored Smoothing (PAS)}
Reading the pose as the $\arg\max$ of the belief ties it to the grid resolution.
PAS refines the position to sub-grid precision and keeps the discrete orientation.
Let $(\mathbf{x}^\star,\theta^\star)=\arg\max_{\mathbf{x},\theta} b_t(\mathbf{x},\theta)$ be the peak cell.
Over a square window $W$ of radius $r$ around $\mathbf{x}^\star$, we marginalize the belief over orientation and take the weighted mean of the positions,
\begin{equation}
\hat{\mathbf{x}} = \frac{\sum_{\mathbf{x}\in W} \big(\sum_{\theta}\exp b_t(\mathbf{x},\theta)\big)\,\mathbf{x}}{\sum_{\mathbf{x}\in W} \sum_{\theta}\exp b_t(\mathbf{x},\theta)} .
\end{equation}
The pose is then $(\hat{\mathbf{x}},\theta^\star)$, shown in Figure~\ref{fig:pas}.
The window keeps far-away mass from biasing the mean, unlike a global soft-argmax.
Its radius $r$ balances accuracy against over-smoothing.

\begin{table}[t]
\centering
\setlength{\tabcolsep}{4pt}
\begin{tabular}{@{}clcccccc@{}}
\toprule[1.5pt]
& & \multicolumn{3}{c}{CV-FSS} & \multicolumn{3}{c}{CV-RHO} \\
\cmidrule(lr){3-5} \cmidrule(lr){6-8}
& Method & 1$\uparrow$ & 3$\uparrow$ & 5$\uparrow$ & 1$\uparrow$ & 3$\uparrow$ & 5$\uparrow$ \\
\midrule
\multirow{6}{*}{\rotatebox[origin=c]{90}{Pos.\ (m)}} & OrienterNet & 0.3 & 3.3 & 8.3 & 7.8 & 31.8 & 43.8 \\
& \textbf{+ SeqLoc (ours)} & \textbf{3.1} & \textbf{23.0} & \textbf{39.9} & \textbf{20.0} & \textbf{59.6} & \textbf{70.3} \\
\addlinespace[2pt]
& OSMLoc & 0.4 & 3.6 & 8.8 & 8.5 & 33.6 & 46.0 \\
& \textbf{+ SeqLoc (ours)} & \textbf{7.1} & \textbf{32.3} & \textbf{49.8} & \textbf{21.5} & \textbf{62.4} & \textbf{73.5} \\
\addlinespace[2pt]
& RHO & 0.6 & 4.8 & 10.6 & 9.0 & 35.3 & 47.9 \\
& \textbf{+ SeqLoc (ours)} & \textbf{14.5} & \textbf{55.9} & \textbf{75.0} & \textbf{37.6} & \textbf{86.8} & \textbf{94.2} \\
\midrule
\multirow{6}{*}{\rotatebox[origin=c]{90}{Ori.\ ($^\circ$)}} & OrienterNet & 3.6 & 11.4 & 18.0 & 15.0 & 39.3 & 53.6 \\
& \textbf{+ SeqLoc (ours)} & \textbf{24.5} & \textbf{49.4} & \textbf{60.1} & \textbf{39.8} & \textbf{69.0} & \textbf{77.2} \\
\addlinespace[2pt]
& OSMLoc & 5.2 & 14.7 & 22.3 & 19.9 & 46.6 & 58.3 \\
& \textbf{+ SeqLoc (ours)} & \textbf{27.4} & \textbf{54.3} & \textbf{65.3} & \textbf{46.0} & \textbf{77.8} & \textbf{83.6} \\
\addlinespace[2pt]
& RHO & 6.7 & 18.4 & 27.0 & 25.0 & 53.8 & 63.4 \\
& \textbf{+ SeqLoc (ours)} & \textbf{55.3} & \textbf{86.2} & \textbf{92.2} & \textbf{75.0} & \textbf{95.4} & \textbf{97.0} \\
\bottomrule[1.5pt]
\end{tabular}
\caption{Overall results on the two benchmarks, averaged over the five regions of CV-FSS and the seven cities of CV-RHO.
We report position recall (\%) at 1/3/5\,m and orientation recall (\%) at 1/3/5$^\circ$.
``+ SeqLoc (ours)'' denotes the corresponding single-frame method combined with our online test-time sequence aggregation.}
\label{tab:summary}
\end{table}

\begin{table*}[t]
\centering
\setlength{\tabcolsep}{6pt}
{\small
\begin{tabular}{@{}clccccccccccccccc@{}}
\toprule[1.5pt]
& & \multicolumn{3}{c}{Neuanspach} & \multicolumn{3}{c}{Tirol} & \multicolumn{3}{c}{Bad Herrenalb} & \multicolumn{3}{c}{Speicher} & \multicolumn{3}{c}{Overgaard} \\
\cmidrule(lr){3-5} \cmidrule(lr){6-8} \cmidrule(lr){9-11} \cmidrule(lr){12-14} \cmidrule(lr){15-17}
& Method & 1$\uparrow$ & 3$\uparrow$ & 5$\uparrow$ & 1$\uparrow$ & 3$\uparrow$ & 5$\uparrow$ & 1$\uparrow$ & 3$\uparrow$ & 5$\uparrow$ & 1$\uparrow$ & 3$\uparrow$ & 5$\uparrow$ & 1$\uparrow$ & 3$\uparrow$ & 5$\uparrow$ \\
\midrule
\multirow{6}{*}{\rotatebox[origin=c]{90}{Pos.\ (m)}} & OrienterNet & 0.4 & 4.8 & 10.2 & 0.5 & 4.8 & 12.8 & 0.2 & 2.4 & 5.4 & 0.4 & 3.2 & 7.9 & 0.2 & 1.5 & 5.0 \\
& \textbf{~~+ SeqLoc (ours)} & \textbf{4.3} & \textbf{26.6} & \textbf{45.9} & \textbf{4.9} & \textbf{29.6} & \textbf{50.8} & \textbf{0.8} & \textbf{22.7} & \textbf{29.3} & \textbf{3.5} & \textbf{21.6} & \textbf{38.8} & \textbf{1.9} & \textbf{14.6} & \textbf{34.7} \\
\addlinespace[2pt]
& OSMLoc & 0.4 & 5.2 & 10.7 & 0.9 & 5.3 & 15.1 & 0.2 & 2.2 & 4.9 & 0.4 & 3.5 & 8.0 & 0.2 & 1.9 & 5.3 \\
& \textbf{~~+ SeqLoc (ours)} & \textbf{5.6} & \textbf{31.6} & \textbf{55.2} & \textbf{8.8} & \textbf{33.3} & \textbf{66.1} & \textbf{10.0} & \textbf{38.5} & \textbf{44.4} & \textbf{5.2} & \textbf{35.7} & \textbf{40.0} & \textbf{5.8} & \textbf{22.6} & \textbf{43.3} \\
\addlinespace[2pt]
& RHO & 0.5 & 5.7 & 12.6 & 1.6 & 9.2 & 21.0 & 0.2 & 2.0 & 4.8 & 0.5 & 4.0 & 8.3 & 0.2 & 2.9 & 6.3 \\
& \textbf{~~+ SeqLoc (ours)} & \textbf{16.4} & \textbf{60.6} & \textbf{84.0} & \textbf{15.5} & \textbf{58.0} & \textbf{78.8} & \textbf{11.4} & \textbf{45.0} & \textbf{58.6} & \textbf{13.5} & \textbf{57.6} & \textbf{78.2} & \textbf{15.9} & \textbf{58.2} & \textbf{75.6} \\
\midrule
\multirow{6}{*}{\rotatebox[origin=c]{90}{Ori.\ ($^\circ$)}} & OrienterNet & 5.2 & 15.4 & 22.4 & 5.0 & 14.0 & 25.1 & 2.2 & 10.2 & 14.9 & 3.5 & 11.3 & 17.4 & 2.3 & 6.0 & 10.2 \\
& \textbf{~~+ SeqLoc (ours)} & \textbf{26.9} & \textbf{58.5} & \textbf{67.7} & \textbf{32.4} & \textbf{60.0} & \textbf{71.1} & \textbf{27.7} & \textbf{50.6} & \textbf{57.6} & \textbf{21.2} & \textbf{45.7} & \textbf{57.3} & \textbf{14.5} & \textbf{32.1} & \textbf{46.9} \\
\addlinespace[2pt]
& OSMLoc & 6.5 & 19.2 & 27.6 & 8.6 & 22.7 & 36.4 & 3.1 & 10.7 & 16.2 & 3.9 & 10.9 & 16.4 & 3.9 & 9.8 & 14.7 \\
& \textbf{~~+ SeqLoc (ours)} & \textbf{29.7} & \textbf{60.5} & \textbf{75.0} & \textbf{34.8} & \textbf{71.1} & \textbf{80.9} & \textbf{30.6} & \textbf{53.7} & \textbf{60.8} & \textbf{24.4} & \textbf{48.7} & \textbf{59.4} & \textbf{17.7} & \textbf{37.3} & \textbf{50.4} \\
\addlinespace[2pt]
& RHO & 8.2 & 24.1 & 33.5 & 11.4 & 32.3 & 48.9 & 4.2 & 11.0 & 17.3 & 4.3 & 10.6 & 15.5 & 5.3 & 14.0 & 19.8 \\
& \textbf{~~+ SeqLoc (ours)} & \textbf{63.8} & \textbf{93.4} & \textbf{96.8} & \textbf{63.7} & \textbf{91.2} & \textbf{94.6} & \textbf{45.8} & \textbf{80.3} & \textbf{91.0} & \textbf{56.0} & \textbf{85.8} & \textbf{92.0} & \textbf{47.1} & \textbf{80.3} & \textbf{86.5} \\
\bottomrule[1.5pt]
\end{tabular}}
\caption{Per-region results on CV-FSS.
Regardless of the backbone, our SeqLoc method improves the performance by a large margin on every region. ``Pos.'' denotes Position (meter), while ``Ori.'' denotes Orientation (degree).}
\label{tab:cvfss}
\end{table*}

\begin{table*}[t]
\centering
\setlength{\tabcolsep}{2pt}
{\small
\begin{tabular}{@{}clccccccccccccccccccccc@{}}
\toprule[1.5pt]
& & \multicolumn{3}{c}{Berlin} & \multicolumn{3}{c}{Chicago} & \multicolumn{3}{c}{Detroit} & \multicolumn{3}{c}{Montrouge} & \multicolumn{3}{c}{San Francisco} & \multicolumn{3}{c}{Toulouse} & \multicolumn{3}{c}{Washington} \\
\cmidrule(lr){3-5} \cmidrule(lr){6-8} \cmidrule(lr){9-11} \cmidrule(lr){12-14} \cmidrule(lr){15-17} \cmidrule(lr){18-20} \cmidrule(lr){21-23}
& Method & 1$\uparrow$ & 3$\uparrow$ & 5$\uparrow$ & 1$\uparrow$ & 3$\uparrow$ & 5$\uparrow$ & 1$\uparrow$ & 3$\uparrow$ & 5$\uparrow$ & 1$\uparrow$ & 3$\uparrow$ & 5$\uparrow$ & 1$\uparrow$ & 3$\uparrow$ & 5$\uparrow$ & 1$\uparrow$ & 3$\uparrow$ & 5$\uparrow$ & 1$\uparrow$ & 3$\uparrow$ & 5$\uparrow$ \\
\midrule
\multirow{6}{*}{\rotatebox[origin=c]{90}{Pos.\ (m)}} & OrienterNet & 2.3 & 18.2 & 33.5 & 4.5 & 25.5 & 37.2 & 18.3 & 56.2 & 65.4 & 6.7 & 27.0 & 36.5 & 7.2 & 35.7 & 50.3 & 9.5 & 30.5 & 42.3 & 6.4 & 29.8 & 41.4 \\
& \textbf{~~+ SeqLoc (ours)} & \textbf{6.0} & \textbf{51.2} & \textbf{69.5} & \textbf{18.6} & \textbf{57.7} & \textbf{67.3} & \textbf{35.9} & \textbf{76.7} & \textbf{82.6} & \textbf{17.5} & \textbf{53.2} & \textbf{59.8} & \textbf{18.5} & \textbf{65.5} & \textbf{76.3} & \textbf{25.5} & \textbf{57.1} & \textbf{71.2} & \textbf{17.8} & \textbf{55.7} & \textbf{65.5} \\
\addlinespace[2pt]
& OSMLoc & 2.4 & 22.0 & 39.7 & 5.9 & 27.4 & 40.0 & 19.8 & 59.0 & 68.1 & 8.0 & 29.3 & 38.2 & 6.2 & 32.2 & 47.2 & 10.7 & 34.2 & 46.1 & 6.5 & 31.4 & 42.7 \\
& \textbf{~~+ SeqLoc (ours)} & \textbf{6.5} & \textbf{55.7} & \textbf{77.5} & \textbf{21.1} & \textbf{60.6} & \textbf{70.1} & \textbf{38.5} & \textbf{79.9} & \textbf{85.3} & \textbf{19.2} & \textbf{55.8} & \textbf{62.2} & \textbf{19.0} & \textbf{66.5} & \textbf{77.1} & \textbf{27.5} & \textbf{60.5} & \textbf{75.0} & \textbf{18.7} & \textbf{57.8} & \textbf{67.5} \\
\addlinespace[2pt]
& RHO & 2.4 & 26.2 & 45.5 & 7.1 & 29.3 & 43.1 & 20.9 & 61.6 & 70.9 & 9.0 & 31.5 & 40.3 & 5.1 & 27.5 & 42.0 & 11.7 & 38.1 & 49.5 & 6.6 & 32.9 & 44.1 \\
& \textbf{~~+ SeqLoc (ours)} & \textbf{12.1} & \textbf{73.9} & \textbf{95.0} & \textbf{32.0} & \textbf{81.4} & \textbf{89.5} & \textbf{54.3} & \textbf{96.0} & \textbf{98.3} & \textbf{40.6} & \textbf{91.2} & \textbf{93.7} & \textbf{32.3} & \textbf{85.4} & \textbf{92.5} & \textbf{54.1} & \textbf{86.6} & \textbf{94.0} & \textbf{38.1} & \textbf{93.1} & \textbf{96.3} \\
\midrule
\multirow{6}{*}{\rotatebox[origin=c]{90}{Ori.\ ($^\circ$)}} & OrienterNet & 9.9 & 29.2 & 43.6 & 13.0 & 34.6 & 49.3 & 23.3 & 55.9 & 71.5 & 18.0 & 42.2 & 54.3 & 12.3 & 36.6 & 55.0 & 15.4 & 39.5 & 51.0 & 13.2 & 36.8 & 50.8 \\
& \textbf{~~+ SeqLoc (ours)} & \textbf{39.8} & \textbf{68.4} & \textbf{76.1} & \textbf{42.4} & \textbf{70.9} & \textbf{76.8} & \textbf{50.4} & \textbf{78.3} & \textbf{84.8} & \textbf{37.2} & \textbf{64.2} & \textbf{70.6} & \textbf{29.5} & \textbf{66.3} & \textbf{80.8} & \textbf{41.4} & \textbf{71.7} & \textbf{77.7} & \textbf{37.8} & \textbf{63.4} & \textbf{73.9} \\
\addlinespace[2pt]
& OSMLoc & 15.5 & 40.0 & 51.6 & 22.1 & 45.7 & 56.5 & 26.8 & 62.3 & 75.5 & 21.0 & 46.3 & 57.0 & 18.7 & 45.5 & 57.5 & 19.1 & 44.7 & 56.6 & 15.8 & 41.4 & 53.6 \\
& \textbf{~~+ SeqLoc (ours)} & \textbf{46.6} & \textbf{80.0} & \textbf{86.1} & \textbf{51.8} & \textbf{84.8} & \textbf{85.3} & \textbf{55.8} & \textbf{85.8} & \textbf{90.4} & \textbf{41.7} & \textbf{70.7} & \textbf{75.4} & \textbf{36.9} & \textbf{75.5} & \textbf{85.6} & \textbf{46.1} & \textbf{78.8} & \textbf{83.8} & \textbf{42.9} & \textbf{69.3} & \textbf{78.3} \\
\addlinespace[2pt]
& RHO & 21.8 & 49.9 & 61.7 & 29.6 & 57.3 & 63.4 & 31.1 & 68.0 & 79.8 & 23.4 & 50.3 & 60.3 & 27.6 & 54.3 & 59.6 & 22.9 & 50.9 & 61.8 & 18.7 & 45.9 & 57.2 \\
& \textbf{~~+ SeqLoc (ours)} & \textbf{75.4} & \textbf{95.9} & \textbf{97.5} & \textbf{76.9} & \textbf{93.9} & \textbf{95.0} & \textbf{81.6} & \textbf{98.0} & \textbf{98.9} & \textbf{71.9} & \textbf{94.0} & \textbf{95.5} & \textbf{73.2} & \textbf{95.8} & \textbf{97.0} & \textbf{72.0} & \textbf{94.5} & \textbf{97.1} & \textbf{74.1} & \textbf{95.9} & \textbf{97.9} \\
\bottomrule[1.5pt]
\end{tabular}}
\caption{Per-city results on CV-RHO.
Even on these feature-rich urban scenes, SeqLoc improves every city and boosts recall significantly. ``Pos.'' denotes Position (meter), while ``Ori.'' denotes Orientation (degree).}
\label{tab:cvrho}
\end{table*}

\section{Experiments}
\subsection{Implementation Details}
To study the effectiveness of SeqLoc in both feature-sparse and feature-rich scenes, we evaluate it on CV-FSS and CV-RHO~\cite{zheng2026rho} using three representative OSM-based metric CVGL backbones, namely OrienterNet~\cite{sarlin2023orienternet}, OSMLoc~\cite{liao2026osmloc}, and RHO~\cite{zheng2026rho}.
We focus on the OSM-based metric CVGL because OSM is more lightweight, easier to acquire, and updated more actively compared to satellite images.
Note that SeqLoc is \textbf{backbone-agnostic}, meaning that it can be applied to any OSM-based metric CVGL backbone as long as the CVGL model outputs a pose likelihood volume.
All backbones are trained from scratch on each scene, and SeqLoc then operates on the likelihood volumes they output without updating any weights.
These volumes are discretized at $0.5$\,m per cell with $256$ orientation bins.
We fix the MGR weight $\varepsilon = 0.01$ and the PAS radius $r = 3$ for all scenes and backbones.
SeqLoc is deterministic, so every number comes from a single run on one 40\,GB A100 GPU under Linux.

\subsection{Quantitative Results}
We report position recall at $1/3/5$\,m and orientation recall at $1/3/5^\circ$, defined as the percentage of frames whose pose error falls within each bound.
These are commonly used metrics for metric CVGL~\cite{sarlin2023orienternet,zheng2026rho}.
Table~\ref{tab:summary} reports the averages over the two benchmarks, CV-FSS and CV-RHO, for all backbones.
On every backbone and benchmark, SeqLoc improves position and orientation recall by a large margin, and the improvement is largest on the feature-sparse CV-FSS, where single-frame localization almost fails.
With RHO, position recall at $5$\,m rises from $10.6\%$ to $75.0\%$ and orientation recall at $3^\circ$ from $18.4\%$ to $86.2\%$.
OrienterNet and OSMLoc follow the same trend, which confirms that SeqLoc is effective and backbone-agnostic.

Table~\ref{tab:cvfss} presents the per-region results on CV-FSS.
On every region, single-frame position recall at the strict threshold is near zero, because the sparse observations make the pose ambiguous.
SeqLoc consistently improves recall across all five regions and all backbones.
This consistency shows that the gain comes from the aggregation itself rather than from a particular scene or backbone.

Table~\ref{tab:cvrho} reports the urban CV-RHO, where single-frame localization already performs well due to richer cues.
SeqLoc still improves every city and boosts recall significantly, reaching an average position recall of $86.8\%$ at $3$\,m and orientation recall of $97.0\%$ at $5^\circ$ with RHO.
SeqLoc therefore improves localization on every scene we evaluate, whether the single-frame method fails or already performs well.

\begin{figure}[t]
\centering
\includegraphics[width=\columnwidth]{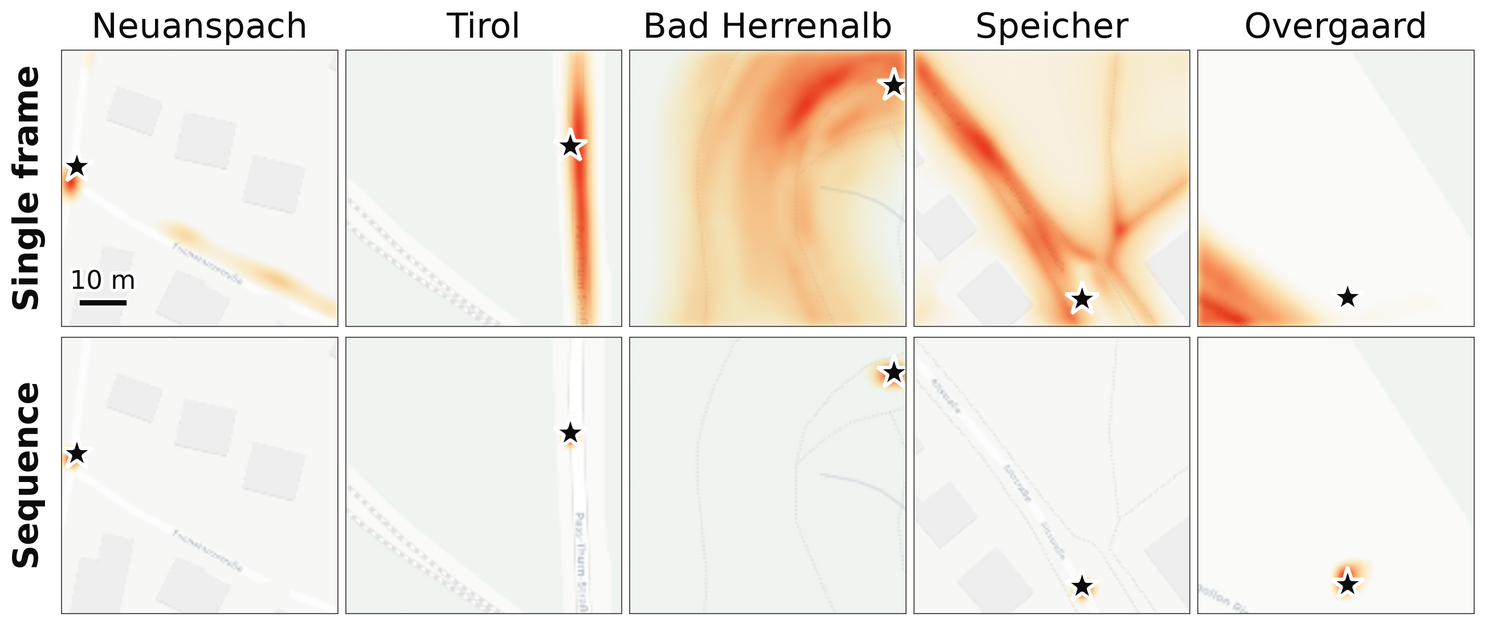}
\caption{Qualitative results on the five regions of CV-FSS.
The pose likelihood is shown in red over OSM, and the star marks the ground-truth pose.
\textbf{Top}: the single-frame likelihood is diffuse and spreads along the road.
\textbf{Bottom}: SeqLoc collapses it into a sharp peak at the true pose.
Zoom in for a better view.}
\label{fig:qualitative}
\end{figure}

\subsection{Qualitative Results}
Figure~\ref{fig:qualitative} shows the pose likelihood before and after SeqLoc on one frame per region of CV-FSS with RHO.
For a single frame, a sparse scene looks nearly the same along the road, so the likelihood forms a ridge whose peak often lies far from the ground truth.
SeqLoc aggregates the likelihoods along the route and collapses this ridge into a sharp peak at the true pose.
This pattern holds across all five regions and matches the large recall gains in Table~\ref{tab:cvfss}.

\begin{table}[t]
\centering
\setlength{\tabcolsep}{4pt}
\begin{tabular}{@{}ccccccccc@{}}
\toprule[1.5pt]
& & & \multicolumn{3}{c}{Pos.\ (m)} & \multicolumn{3}{c}{Ori.\ ($^\circ$)} \\
\cmidrule(lr){4-6} \cmidrule(lr){7-9}
MGR & ETU & PAS & 1$\uparrow$ & 3$\uparrow$ & 5$\uparrow$ & 1$\uparrow$ & 3$\uparrow$ & 5$\uparrow$ \\
\midrule
\multicolumn{3}{c}{Single frame} & 0.2 & 2.9 & 6.3 & 5.3 & 14.0 & 19.8 \\
\addlinespace[2pt]
$\times$ & $\times$ & $\times$ & 3.3 & 17.5 & 27.5 & 10.2 & 30.8 & 37.5 \\
\checkmark & $\times$ & $\times$ & 7.9 & 31.8 & 48.5 & 26.4 & 59.7 & 72.9 \\
\checkmark & \checkmark & $\times$ & 13.3 & 56.3 & 73.8 & 47.1 & 80.3 & 86.5 \\
\textbf{\checkmark} & \textbf{\checkmark} & \textbf{\checkmark} & \textbf{15.9} & \textbf{58.2} & \textbf{75.6} & \textbf{47.1} & \textbf{80.3} & \textbf{86.5} \\
\bottomrule[1.5pt]
\end{tabular}
\caption{Ablation of the three components of SeqLoc with RHO on the Overgaard region of CV-FSS.}
\label{tab:ablation}
\end{table}

\begin{table}[t]
\centering
\setlength{\tabcolsep}{8pt}
\begin{tabular}{@{}ccccccc@{}}
\toprule[1.5pt]
& \multicolumn{3}{c}{Pos.\ (m)} & \multicolumn{3}{c}{Ori.\ ($^\circ$)} \\
\cmidrule(lr){2-4} \cmidrule(lr){5-7}
$\varepsilon$ & 1$\uparrow$ & 3$\uparrow$ & 5$\uparrow$ & 1$\uparrow$ & 3$\uparrow$ & 5$\uparrow$ \\
\midrule
\textbf{0.01} & \textbf{15.9} & \textbf{58.2} & \textbf{75.6} & \textbf{47.1} & \textbf{80.3} & \textbf{86.5} \\
0.02 & 15.3 & 56.9 & 74.5 & 45.7 & 79.4 & 86.0 \\
0.03 & 15.6 & 56.4 & 74.1 & 45.7 & 79.3 & 86.3 \\
0.04 & 15.5 & 56.3 & 73.9 & 45.7 & 79.0 & 86.4 \\
0.05 & 15.2 & 55.8 & 73.5 & 45.4 & 78.9 & 86.4 \\
\bottomrule[1.5pt]
\end{tabular}
\caption{Effect of the recovery weight $\varepsilon$ of MGR with RHO on the Overgaard region of CV-FSS.}
\label{tab:mgr}
\end{table}

\subsection{Ablation Studies}
\subsubsection{Component Contributions}
Table~\ref{tab:ablation} reports a cumulative ablation that enables the three components successively over a naive recursive fusion of the per-frame likelihood volumes.
Recursive fusion alone already yields a large gain over the single-frame method, confirming that the online test-time aggregation is beneficial, and adding MGR then ETU produces two further large improvements as they respectively restore wrongly suppressed poses and down-weight high-entropy likelihood volumes, with ETU contributing the most to the position gain.
Finally, PAS sharpens the position estimate to sub-grid precision and leaves orientation recall unchanged, since it operates only on the position component of the pose.

\subsubsection{Recovery Weight of MGR}
Table~\ref{tab:mgr} studies the recovery weight $\varepsilon$ that controls how much map-shaped mass MGR injects into the belief at each step.
This weight balances two opposing requirements.
It must stay large enough for a wrongly suppressed pose to keep a floor of probability and recover, yet small enough not to overwrite the likelihood already accumulated in the belief.
Increasing $\varepsilon$ from $0.01$ to $0.05$ lowers recall on all six metrics, because a heavier recovery term pulls the belief toward the generic road prior and blurs its accumulated peak.
Since $\varepsilon=0.01$ gives the best result throughout, we fix it for all scenes and backbones.

\subsubsection{Window Radius of PAS}
Table~\ref{tab:pas} explores the effect of the PAS window radius $r$ on a feature-sparse scene and an urban scene that differ sharply in their sensitivity to $r$.
At the strict $1$\,m threshold, the most demanding position metric, recall on rural Overgaard rises with $r$, peaks at a moderate radius, and then declines once a larger window over-smooths the estimate, whereas on urban Berlin it stays almost constant over the full range.
We attribute this to the width of the pose belief, a broad ridge along the road when the scene is sparse but a sharp peak when it is structure-rich, so a larger averaging window helps only in the former case.
Since the sparse scene turns over at a moderate radius, we fix a small conservative $r=3$ for all scenes and backbones, comfortably below this over-smoothing onset, rather than tuning it per scene.

\begin{table}[t]
\centering
\setlength{\tabcolsep}{10pt}
\begin{tabular}{@{}ccccccc@{}}
\toprule[1.5pt]
& \multicolumn{3}{c}{Overgaard} & \multicolumn{3}{c}{Berlin} \\
\cmidrule(lr){2-4} \cmidrule(lr){5-7}
$r$ & 1$\uparrow$ & 3$\uparrow$ & 5$\uparrow$ & 1$\uparrow$ & 3$\uparrow$ & 5$\uparrow$ \\
\midrule
1 & 14.1 & 56.7 & 74.8 & 12.0 & 72.6 & 94.6 \\
2 & 14.8 & 57.7 & 75.3 & 12.1 & 72.8 & 94.8 \\
3 & 15.9 & 58.2 & 75.6 & 12.1 & 73.9 & 95.0 \\
4 & 16.2 & 58.7 & 75.9 & 12.1 & 74.4 & 95.2 \\
5 & 17.2 & 59.2 & 76.0 & 12.6 & 74.7 & 95.2 \\
6 & 17.9 & 59.1 & 76.1 & 12.7 & 74.6 & 95.2 \\
7 & 17.4 & 59.5 & 76.4 & 12.8 & 75.0 & 95.4 \\
8 & 17.5 & 59.8 & 76.6 & 12.9 & 75.2 & 95.4 \\
9 & 17.3 & 59.8 & 76.6 & 12.9 & 75.4 & 95.4 \\
10 & 16.7 & 59.9 & 76.8 & 12.9 & 75.4 & 95.6 \\
\bottomrule[1.5pt]
\end{tabular}
\caption{Effect of the window radius $r$ of PAS with RHO on the feature-sparse Overgaard region of CV-FSS and the urban Berlin city of CV-RHO.
PAS refines only the position, so orientation recall is unaffected and omitted here.}
\label{tab:pas}
\end{table}

\section{Conclusion}
In this work, we look into feature-sparse OSM-based metric Cross-View Geo-Localization and establish the CV-FSS benchmark, which pairs sequential panoramas from five rural regions with aligned OSM tiles. 
We put forward SeqLoc, an online test-time sequence aggregation solution which recursively holds a log-belief volume.  
Comprehensive experiments on the proposed CV-FSS and public CV-RHO benchmarks demonstrate that SeqLoc greatly outstrips single-frame localization.

In the future, we intend to investigate challenging Cross-View Geo-Localization under large off-nadir views with severe perspective distortions and frequent occlusions.

\bibliography{references}

@article{chalvatzaras2022survey,
  title={A survey on map-based localization techniques for autonomous vehicles},
  author={Chalvatzaras, Athanasios and Pratikakis, Ioannis and Amanatiadis, Angelos A},
  journal={IEEE Transactions on intelligent vehicles},
  volume={8},
  number={2},
  pages={1574--1596},
  year={2022},
  publisher={IEEE}
}

@article{ullah2024mobile,
  title={Mobile robot localization: Current challenges and future prospective},
  author={Ullah, Inam and Adhikari, Deepak and Khan, Habib and Anwar, M Shahid and Ahmad, Shabir and Bai, Xiaoshan},
  journal={Computer Science Review},
  volume={53},
  pages={100651},
  year={2024},
  publisher={Elsevier}
}

@inproceedings{zheng2025scene,
  title={Scene-agnostic pose regression for visual localization},
  author={Zheng, Junwei and Liu, Ruiping and Chen, Yufan and Chen, Zhenfang and Yang, Kailun and Zhang, Jiaming and Stiefelhagen, Rainer},
  booktitle={Proceedings of the IEEE/CVF Conference on Computer Vision and Pattern Recognition},
  pages={27092--27102},
  year={2025}
}

@inproceedings{sarlin2023orienternet,
  title={Orienternet: Visual localization in 2d public maps with neural matching},
  author={Sarlin, Paul-Edouard and DeTone, Daniel and Yang, Tsun-Yi and Avetisyan, Armen and Straub, Julian and Malisiewicz, Tomasz and Bulo, Samuel Rota and Newcombe, Richard and Kontschieder, Peter and Balntas, Vasileios},
  booktitle={Proceedings of the IEEE/CVF Conference on Computer Vision and Pattern Recognition},
  pages={21632--21642},
  year={2023}
}

@inproceedings{zheng2026rho,
  title={RHO: Robust Holistic OSM-Based Metric Cross-View Geo-Localization},
  author={Zheng, Junwei and Dai, Ruize and Liu, Ruiping and Zeng, Zichao and Chen, Yufan and Wang, Fangjinhua and Peng, Kunyu and Yang, Kailun and Zhang, Jiaming and Stiefelhagen, Rainer},
  booktitle={Proceedings of the IEEE/CVF Conference on Computer Vision and Pattern Recognition},
  pages={33727--33737},
  year={2026}
}

@article{thrun2001robust,
  title={Robust Monte Carlo localization for mobile robots},
  author={Thrun, Sebastian and Fox, Dieter and Burgard, Wolfram and Dellaert, Frank},
  journal={Artificial intelligence},
  volume={128},
  number={1-2},
  pages={99--141},
  year={2001},
  publisher={Elsevier}
}

@article{anguelov2010google,
  title={Google street view: Capturing the world at street level},
  author={Anguelov, Dragomir and Dulong, Carole and Filip, Daniel and Frueh, Christian and Lafon, St{\'e}phane and Lyon, Richard and Ogale, Abhijit and Vincent, Luc and Weaver, Josh},
  journal={Computer},
  volume={43},
  number={6},
  pages={32--38},
  year={2010},
  publisher={IEEE}
}

@inproceedings{hu2018cvm,
  title={Cvm-net: Cross-view matching network for image-based ground-to-aerial geo-localization},
  author={Hu, Sixing and Feng, Mengdan and Nguyen, Rang MH and Lee, Gim Hee},
  booktitle={Proceedings of the IEEE conference on computer vision and pattern recognition},
  pages={7258--7267},
  year={2018}
}

@inproceedings{zhu2022transgeo,
  title={TransGeo: Transformer Is All You Need for Cross-view Image Geo-localization},
  author={Zhu, Sijie and Shah, Mubarak and Chen, Chen},
  booktitle={CVPR},
  year={2022}
}

@inproceedings{deuser2023sample4geo,
  title={Sample4Geo: Hard Negative Sampling For Cross-View Geo-Localisation},
  author={Deuser, Fabian and Habel, Konrad and Oswald, Norbert},
  booktitle={ICCV},
  year={2023}
}

@inproceedings{fervers2024statewide,
  title={Statewide visual geolocalization in the wild},
  author={Fervers, Florian and Bullinger, Sebastian and Bodensteiner, Christoph and Arens, Michael and Stiefelhagen, Rainer},
  booktitle={ECCV},
  year={2024}
}

@inproceedings{ye2025where,
  title={Where am I? Cross-view geo-localization with natural language descriptions},
  author={Ye, Junyan and Lin, Honglin and Ou, Leyan and Chen, Dairong and Wang, Zihao and Zhu, Qi and He, Conghui and Li, Weijia},
  booktitle={CVPR},
  year={2025}
}

@inproceedings{liu2019lending,
  title={Lending orientation to neural networks for cross-view geo-localization},
  author={Liu, Liu and Li, Hongdong},
  booktitle={Proceedings of the IEEE/CVF conference on computer vision and pattern recognition},
  pages={5624--5633},
  year={2019}
}

@article{shi2019spatial,
  title={Spatial-aware feature aggregation for image based cross-view geo-localization},
  author={Shi, Yujiao and Liu, Liu and Yu, Xin and Li, Hongdong},
  journal={Advances in Neural Information Processing Systems},
  volume={32},
  year={2019}
}

@article{yang2021cross,
  title={Cross-view geo-localization with layer-to-layer transformer},
  author={Yang, Hongji and Lu, Xiufan and Zhu, Yingying},
  journal={Advances in Neural Information Processing Systems},
  volume={34},
  pages={29009--29020},
  year={2021}
}

@inproceedings{zhu2021vigor,
  title={Vigor: Cross-view image geo-localization beyond one-to-one retrieval},
  author={Zhu, Sijie and Yang, Taojiannan and Chen, Chen},
  booktitle={Proceedings of the IEEE/CVF Conference on Computer Vision and Pattern Recognition},
  pages={3640--3649},
  year={2021}
}

@inproceedings{shi2022beyond,
  title={Beyond cross-view image retrieval: Highly accurate vehicle localization using satellite image},
  author={Shi, Yujiao and Li, Hongdong},
  booktitle={Proceedings of the IEEE/CVF Conference on Computer Vision and Pattern Recognition},
  pages={17010--17020},
  year={2022}
}

@inproceedings{lentsch2023slicematch,
  title={Slicematch: Geometry-guided aggregation for cross-view pose estimation},
  author={Lentsch, Ted and Xia, Zimin and Caesar, Holger and Kooij, Julian FP},
  booktitle={Proceedings of the IEEE/CVF Conference on Computer Vision and Pattern Recognition},
  pages={17225--17234},
  year={2023}
}

@article{wang2023fine,
  title={Fine-grained cross-view geo-localization using a correlation-aware homography estimator},
  author={Wang, Xiaolong and Xu, Runsen and Cui, Zhuofan and Wan, Zeyu and Zhang, Yu},
  journal={Advances in Neural Information Processing Systems},
  volume={36},
  pages={5301--5319},
  year={2023}
}

@article{xia2023convolutional,
  title={Convolutional cross-view pose estimation},
  author={Xia, Zimin and Booij, Olaf and Kooij, Julian FP},
  journal={IEEE Transactions on Pattern Analysis and Machine Intelligence},
  volume={46},
  number={5},
  pages={3813--3831},
  year={2023},
  publisher={IEEE}
}

@inproceedings{xia2025fg,
  title={FG\^{} 2: Fine-Grained Cross-View Localization by Fine-Grained Feature Matching},
  author={Xia, Zimin and Alahi, Alexandre},
  booktitle={Proceedings of the IEEE/CVF Conference on Computer Vision and Pattern Recognition},
  pages={6362--6372},
  year={2025}
}

@inproceedings{xia2022visual,
  title={Visual cross-view metric localization with dense uncertainty estimates},
  author={Xia, Zimin and Booij, Olaf and Manfredi, Marco and Kooij, Julian FP},
  booktitle={European Conference on Computer Vision},
  pages={90--106},
  year={2022},
  organization={Springer}
}

@inproceedings{fervers2023uncertainty,
  title={Uncertainty-aware vision-based metric cross-view geolocalization},
  author={Fervers, Florian and Bullinger, Sebastian and Bodensteiner, Christoph and Arens, Michael and Stiefelhagen, Rainer},
  booktitle={Proceedings of the IEEE/CVF Conference on Computer Vision and Pattern Recognition},
  pages={21621--21631},
  year={2023}
}

@inproceedings{wu2024maplocnet,
  title={Maplocnet: Coarse-to-fine feature registration for visual re-localization in navigation maps},
  author={Wu, Hang and Zhang, Zhenghao and Lin, Siyuan and Mu, Xiangru and Zhao, Qiang and Yang, Ming and Qin, Tong},
  booktitle={2024 IEEE/RSJ International Conference on Intelligent Robots and Systems (IROS)},
  pages={13198--13205},
  year={2024},
  organization={IEEE}
}

@article{liao2026osmloc,
  title={OSMLoc: Single image-based visual localization in OpenStreetMap with fused geometric and semantic guidance},
  author={Liao, Youqi and Chen, Xieyuanli and Kang, Shuhao and Li, Jianping and Dong, Zhen and Fan, Hongchao and Yang, Bisheng},
  journal={Information Fusion},
  pages={104562},
  year={2026},
  publisher={Elsevier}
}

@inproceedings{vyas2022gama,
  title={Gama: Cross-view video geo-localization},
  author={Vyas, Shruti and Chen, Chen and Shah, Mubarak},
  booktitle={European Conference on Computer Vision},
  pages={440--456},
  year={2022},
  organization={Springer}
}

@inproceedings{shi2022cvlnet,
  title={Cvlnet: Cross-view semantic correspondence learning for video-based camera localization},
  author={Shi, Yujiao and Yu, Xin and Wang, Shan and Li, Hongdong},
  booktitle={Asian Conference on Computer Vision},
  pages={123--141},
  year={2022},
  organization={Springer}
}

@inproceedings{zhang2023cross,
  title={Cross-view image sequence geo-localization},
  author={Zhang, Xiaohan and Sultani, Waqas and Wshah, Safwan},
  booktitle={Proceedings of the IEEE/CVF Winter Conference on Applications of Computer Vision},
  pages={2914--2923},
  year={2023}
}

@inproceedings{dellaert1999monte,
  title={Monte carlo localization for mobile robots},
  author={Dellaert, Frank and Fox, Dieter and Burgard, Wolfram and Thrun, Sebastian},
  booktitle={Proceedings 1999 IEEE international conference on robotics and automation (Cat. No. 99CH36288C)},
  volume={2},
  pages={1322--1328},
  year={1999},
  organization={IEEE}
}

@inproceedings{milford2012seqslam,
  title={SeqSLAM: Visual route-based navigation for sunny summer days and stormy winter nights},
  author={Milford, Michael J and Wyeth, Gordon F},
  booktitle={2012 IEEE international conference on robotics and automation},
  pages={1643--1649},
  year={2012},
  organization={IEEE}
}

@article{thrun2002probabilistic,
  title={Probabilistic robotics},
  author={Thrun, Sebastian},
  journal={Communications of the ACM},
  volume={45},
  number={3},
  pages={52--57},
  year={2002},
  publisher={ACM New York, NY, USA}
}

@article{mur2016orb,
  title={ORB-SLAM2: an open-source SLAM system for monocular, stereo and RGB-D cameras},
  author={Mur-Artal, Raul and Tard{\'o}s, Juan D},
  journal={arXiv preprint arXiv:1610.06475},
  year={2016}
}

@article{cadena2016past,
  title={Past, present, and future of simultaneous localization and mapping: Toward the robust-perception age},
  author={Cadena, Cesar and Carlone, Luca and Carrillo, Henry and Latif, Yasir and Scaramuzza, Davide and Neira, Jos{\'e} and Reid, Ian and Leonard, John J},
  journal={IEEE Transactions on robotics},
  volume={32},
  number={6},
  pages={1309--1332},
  year={2016},
  publisher={IEEE}
}

@article{wu2024cross,
  title={Cross-view image set geo-localization},
  author={Wu, Qiong and Xia, Panwang and Yu, Lei and Liu, Yi and Xiong, Mingtao and Zhong, Liheng and Chen, Jingdong and Yang, Ming and Zhang, Yongjun and Wan, Yi},
  journal={arXiv preprint arXiv:2412.18852},
  year={2024}
}

@article{gao2026trajectory,
  title={Trajectory-aware Cross-view Geo-localization with Sequential Observations},
  author={Gao, Tianyi and Lin, Jiayu and Beaulieu, Danielle and Jacobs, Nathan},
  journal={arXiv preprint arXiv:2607.15491},
  year={2026}
}

@inproceedings{regmi2021video,
  title={Video geo-localization employing geo-temporal feature learning and gps trajectory smoothing},
  author={Regmi, Krishna and Shah, Mubarak},
  booktitle={Proceedings of the IEEE/CVF International Conference on Computer Vision},
  pages={12126--12135},
  year={2021}
}

@inproceedings{pillai2024garet,
  title={Garet: cross-view video geolocalization with adapters and auto-regressive transformers},
  author={Pillai, Manu S and Rizve, Mamshad Nayeem and Shah, Mubarak},
  booktitle={European Conference on Computer Vision},
  pages={466--483},
  year={2024},
  organization={Springer}
}

@article{huang2024cv,
  title={Cv-cities: Advancing cross-view geo-localization in global cities},
  author={Huang, Gaoshuang and Zhou, Yang and Zhao, Luying and Gan, Wenjian},
  journal={IEEE Journal of Selected Topics in Applied Earth Observations and Remote Sensing},
  volume={18},
  pages={1592--1606},
  year={2024},
  publisher={IEEE}
}

@inproceedings{cao2023tightly,
  title={Tightly-coupled LiDAR-visual SLAM based on geometric features for mobile agents},
  author={Cao, Ke and Liu, Ruiping and Wang, Ze and Peng, Kunyu and Zhang, Jiaming and Zheng, Junwei and Teng, Zhifeng and Yang, Kailun and Stiefelhagen, Rainer},
  booktitle={2023 IEEE International Conference on Robotics and Biomimetics (ROBIO)},
  pages={1--8},
  year={2023},
  organization={IEEE}
}

@inproceedings{wei2024onebev,
  title={OneBEV: Using one panoramic image for bird’s-eye-view semantic mapping},
  author={Wei, Jiale and Zheng, Junwei and Liu, Ruiping and Hu, Jie and Zhang, Jiaming and Stiefelhagen, Rainer},
  booktitle={Asian Conference on Computer Vision},
  pages={377--393},
  year={2024},
  organization={Springer}
}

@article{wei2026onebev++,
  title={OneBEV++: Towards Unifying Bird's-Eye-View Semantic Mapping with Panoramas},
  author={Wei, Jiale and Teng, Zhifeng and Teng, Fei and Zheng, Junwei and Liu, Ruiping and Chen, Yufan and Hu, Jie and Yang, Kailun and Zhang, Jiaming and Stiefelhagen, Rainer},
  journal={IEEE Transactions on Pattern Analysis and Machine Intelligence},
  year={2026},
  publisher={IEEE}
}

@inproceedings{zheng2024open,
  title={Open panoramic segmentation},
  author={Zheng, Junwei and Liu, Ruiping and Chen, Yufan and Peng, Kunyu and Wu, Chengzhi and Yang, Kailun and Zhang, Jiaming and Stiefelhagen, Rainer},
  booktitle={European Conference on Computer Vision},
  pages={164--182},
  year={2024},
  organization={Springer}
}

@article{hu2024deformable,
  title={Deformable mamba for wide field of view segmentation},
  author={Hu, Jie and Zheng, Junwei and Wei, Jiale and Zhang, Jiaming and Stiefelhagen, Rainer},
  journal={arXiv preprint arXiv:2411.16481},
  year={2024}
}

@inproceedings{fan2026more,
  title={More than the Sum: Panorama-Language Models for Adverse Omni-Scenes},
  author={Fan, Weijia and Liu, Ruiping and Wei, Jiale and Chen, Yufan and Zheng, Junwei and Zeng, Zichao and Zhang, Jiaming and Li, Qiufu and Shen, Linlin and Stiefelhagen, Rainer},
  booktitle={Proceedings of the IEEE/CVF Conference on Computer Vision and Pattern Recognition},
  pages={30874--30884},
  year={2026}
}

\twocolumn[%
  \begin{center}
    {\LARGE\bf Supplementary Material for SeqLoc\par}
  \end{center}
  \vskip 2em%
]

This document provides supplementary material for the main paper, \emph{SeqLoc: Beyond the Single Frame for Cross-View Geo-Localization in Feature-Sparse Scenes}.
We detail the training and evaluation setup of the RHO backbone, compare online and offline sequence aggregation, and test SeqLoc under noisy odometry, where it stays robust across the realistic range of drift. We then extend the per-region entropy analysis to all five regions of CV-FSS, present additional data samples, and discuss the societal impact of the work.

\section{Implementation Details}
Table~\ref{tab:supp_config} lists the full training and evaluation configuration of the RHO backbone.
The configurations for OrienterNet and OSMLoc can be found on our project page.
RHO pairs a ResNet-101 image encoder, initialized from ImageNet, with a VGG-19 map encoder trained from scratch.
Each scene is trained independently, and SeqLoc then operates on the resulting likelihood volumes without updating any weights.
The learning-rate schedule uses factor $0.7$ and patience $2$, and the NLL ground truth is the first view of each panorama.

The split of each scene is spatial rather than random, with the test set forming a connected area that the training data surround, so the two never overlap.
On CV-FSS, the five regions provide $27{,}699$ training and $7{,}084$ test panoramas over $102$ test routes, counting only panoramas whose three perspective views are all present.

The three test-time components of SeqLoc add no learned parameters, and we fix the MGR weight $\varepsilon = 0.01$ and the PAS radius $r = 3$ for all scenes and backbones.

\section{Online versus Offline Sequence Aggregation}
Every backbone produces a per-frame pose likelihood volume that a sequence method aggregates into a trajectory, either offline or online.
Offline aligns all frames to one reference and sums them in a single batch, which uses future frames and localizes only that reference frame.
Online instead updates a running belief recursively from past frames alone, localizing every frame.
SeqLoc is our strengthened online variant, adding three components, ETU, MGR, and PAS, on top of plain recursive fusion.
These two modes are hard to compare on equal terms.
Offline draws on future frames that online never sees, and by default it scores only the reference frame of each sequence rather than every frame.
For a common protocol we force offline to localize every frame as well.
This repeats the whole-sequence batch at every frame, so its cost grows steeply with the route length, whereas online keeps a constant per-frame update.
Table~\ref{tab:supp_ttsa} reports this deliberately forced comparison, averaged over the two benchmarks as in the main paper.
Even with the future-frame advantage, offline still trails online on every backbone and benchmark, and SeqLoc widens the gap further, most on the feature-sparse CV-FSS where the single-frame signal is weakest.

\section{Robustness to Noisy Odometry}
SeqLoc propagates its running log-belief through the relative motion $\Delta_t$ read by odometry, warping the previous belief to the current frame by the inverse transform $T_{\Delta_t}^{-1}$ before it fuses the current likelihood volume.
Offline aggregation relies on the same relative motion, using $\Delta_t$ to align all frames to one reference frame before summing them.
In the experiments of the main paper, $\Delta_t$ is taken from the reference poses of the dataset, which is equivalent to an ideal odometry with zero error.
A real odometry instead returns noisy readings whose error accumulates along the route.
Since SeqLoc is an online method for a moving system, we test whether it remains reliable once $\Delta_t$ is corrupted by accumulated odometry noise.

We perturb $\Delta_t$ with a zero-mean Gaussian noise that follows the KITTI convention, where the translation error scales with the step length as a percentage $\sigma_t$ and the heading error scales with it in degrees per meter as $\sigma_r$.
The noise is added to each incremental motion and then integrated, so the reference trajectory drifts like a real front-end rather than being resampled independently at every frame.
Ground truth is never perturbed and only scores the result.

Table~\ref{tab:supp_odom} reports the SeqLoc error and recall across the five noise levels, and both stay \textbf{stable} as the noise grows.
The mean position error holds near $3.7$ m from the ideal setting through the medium level and reaches only $4.2$ m under poor odometry.
The orientation error stays within $2.6$ to $2.8^\circ$ over the same range.
The recall at the wider thresholds tells the same story.
The position recall at $5$ m falls only from $84.0\%$ to $77.1\%$, and the orientation recall at $3^\circ$ from $93.4\%$ to $91.1\%$.
Only the very poor level moves the numbers clearly.
Its $10\%$ translation error and $0.05^\circ$/m of heading noise push the position error to $5.6$ m and the $5$ m recall down to $61.9\%$.
The map corrects the belief at every frame, so the error is bounded by how well a frame matches the map rather than by the accumulated odometry drift.
The good and medium levels, at $1$ to $2\%$ translation drift, match the accuracy that modern visual, visual-inertial, and LiDAR odometry report in practice.
At this real-world accuracy, SeqLoc is \textbf{barely distinguishable from ideal odometry}.
Moving from the ideal case to medium noise adds only $0.1$ m to the position error and $0.1^\circ$ to the orientation error, and it lowers the $5$ m recall by only about $1\%$.
SeqLoc therefore stays \textbf{robust} across the realistic range of odometry noise.

\section{Per-Region Entropy Analysis}
\begin{figure}[t]
\centering
\includegraphics[width=\columnwidth]{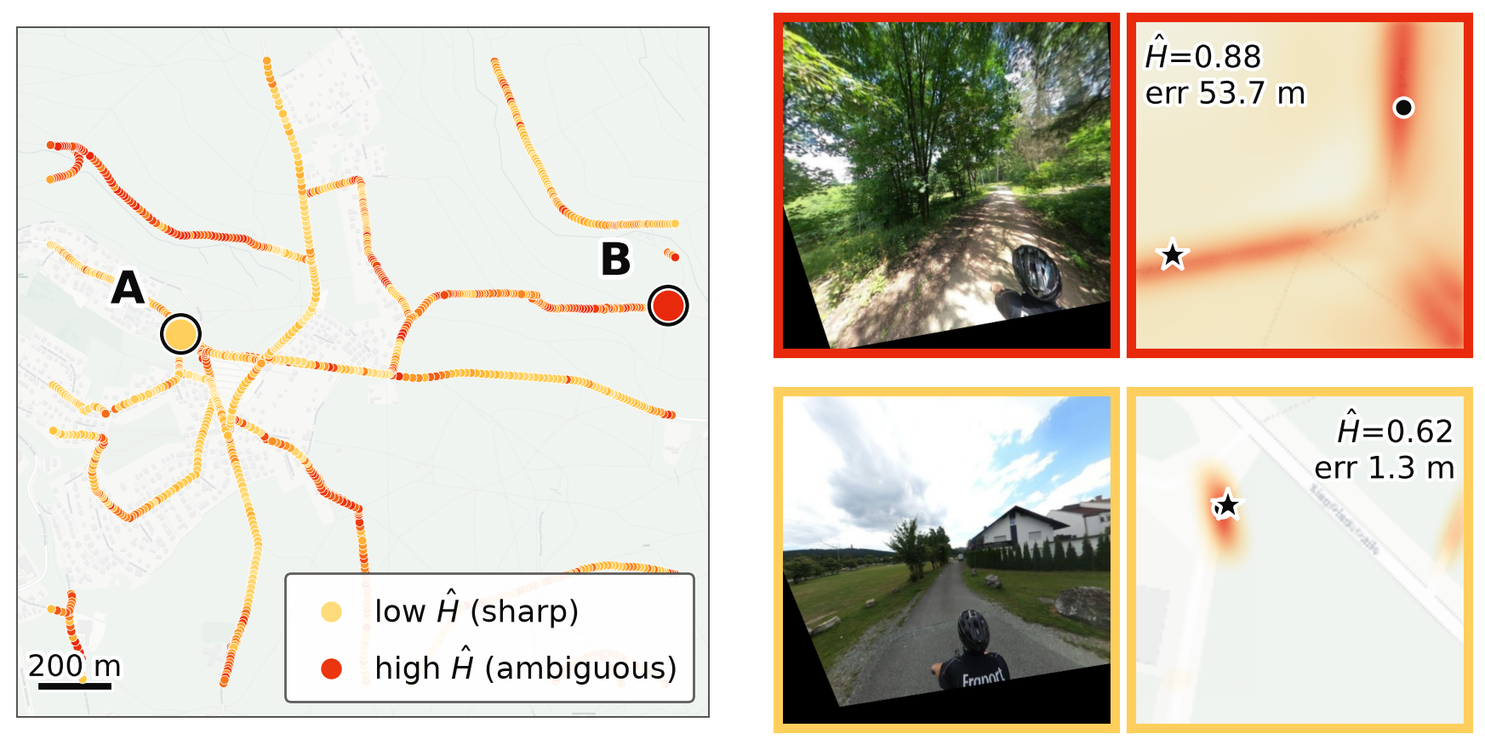}
\caption{Per-region entropy analysis on the Neuanspach region of CV-FSS.
The left map colors each frame of the test route by its normalized entropy $\hat{H}$, from low (sharp, yellow) to high (ambiguous, red).
The right panels contrast a distinctive location A (low $\hat{H}$) with an ambiguous location B (high $\hat{H}$), with black dots at the argmax and stars at the ground truth.}
\label{fig:supp_etu_neuanspach}
\end{figure}

\begin{figure}[t]
\centering
\includegraphics[width=\columnwidth]{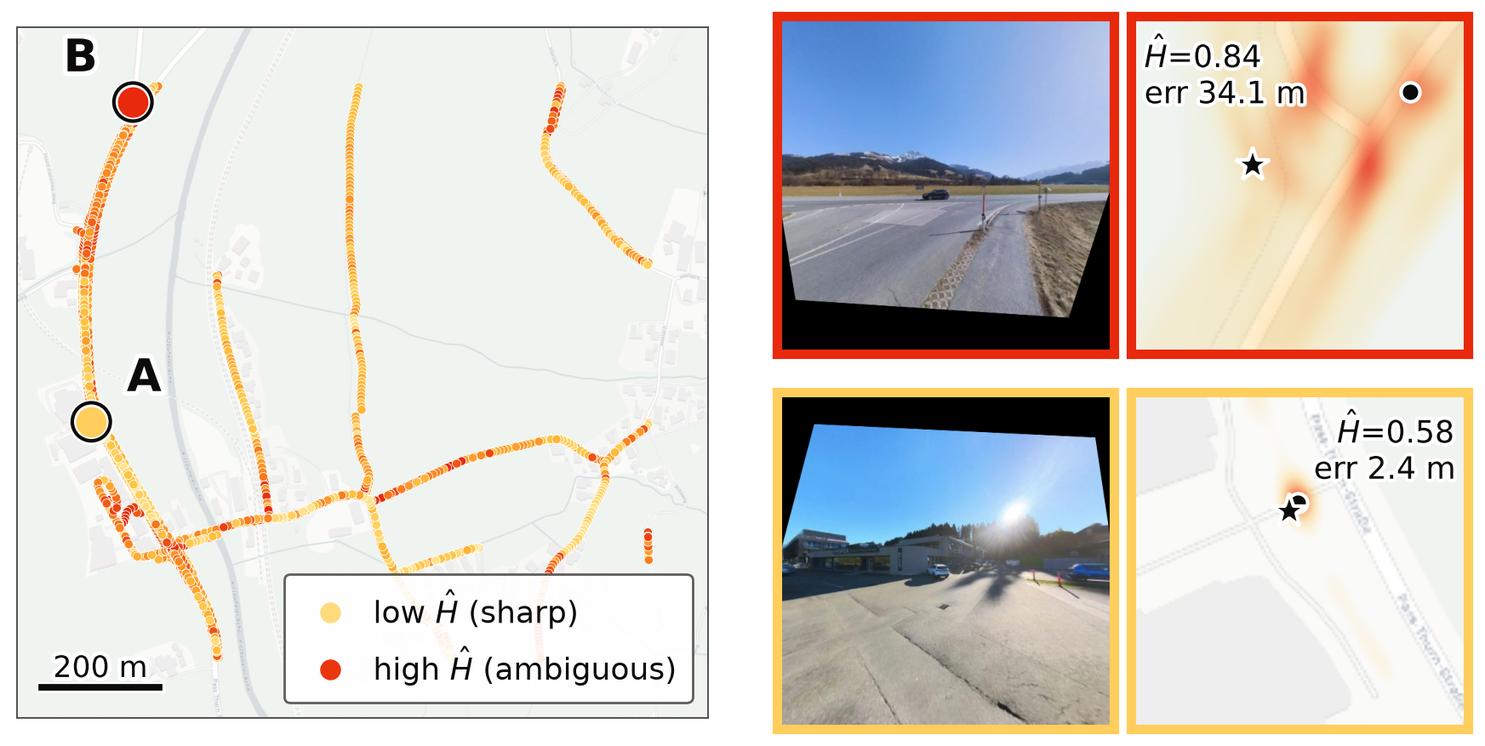}
\caption{Per-region entropy analysis on the Tirol region of CV-FSS, in the same format as Figure~\ref{fig:supp_etu_neuanspach}.}
\label{fig:supp_etu_tirol}
\end{figure}

\begin{figure}[t]
\centering
\includegraphics[width=\columnwidth]{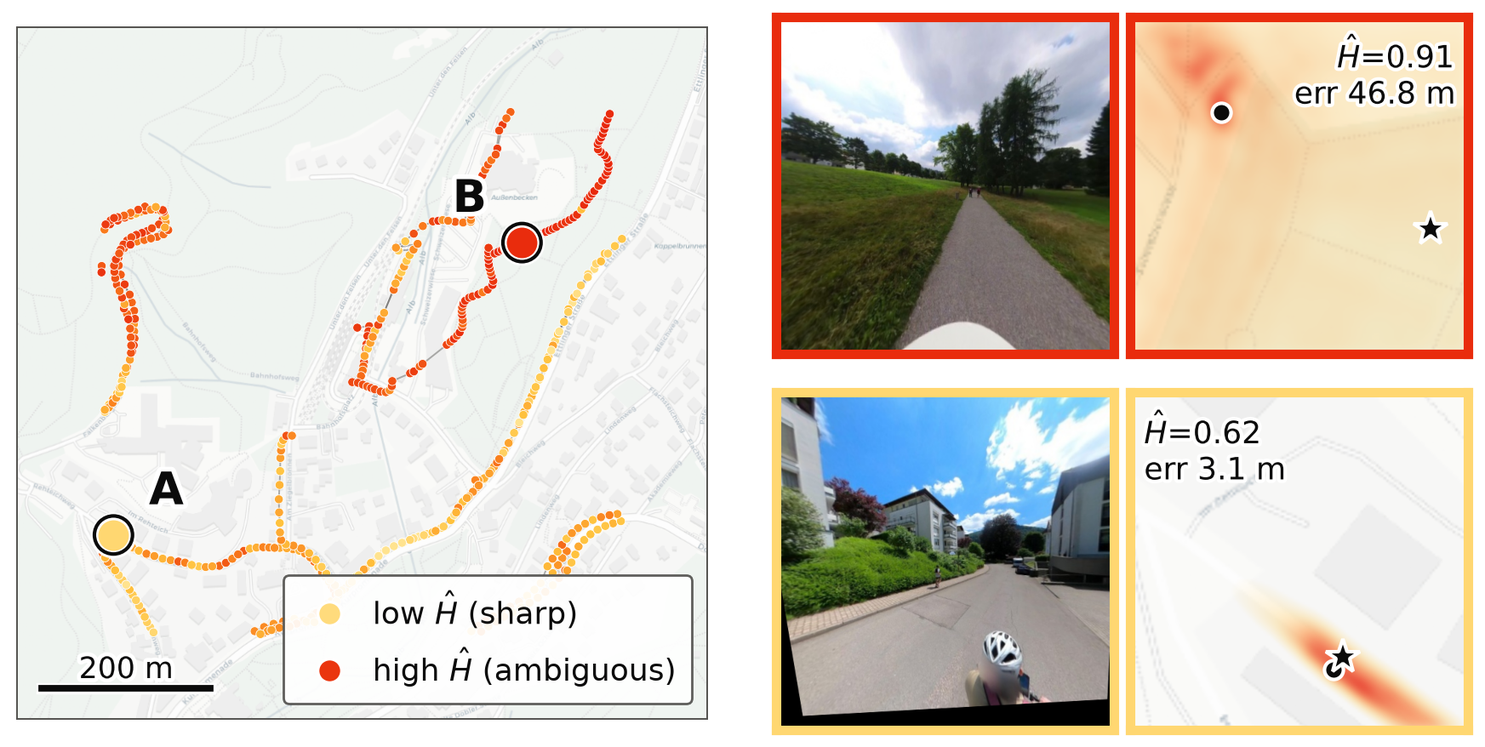}
\caption{Per-region entropy analysis on the Bad Herrenalb region of CV-FSS, in the same format as Figure~\ref{fig:supp_etu_neuanspach}.}
\label{fig:supp_etu_bad_herrenalb}
\end{figure}

\begin{figure}[t]
\centering
\includegraphics[width=\columnwidth]{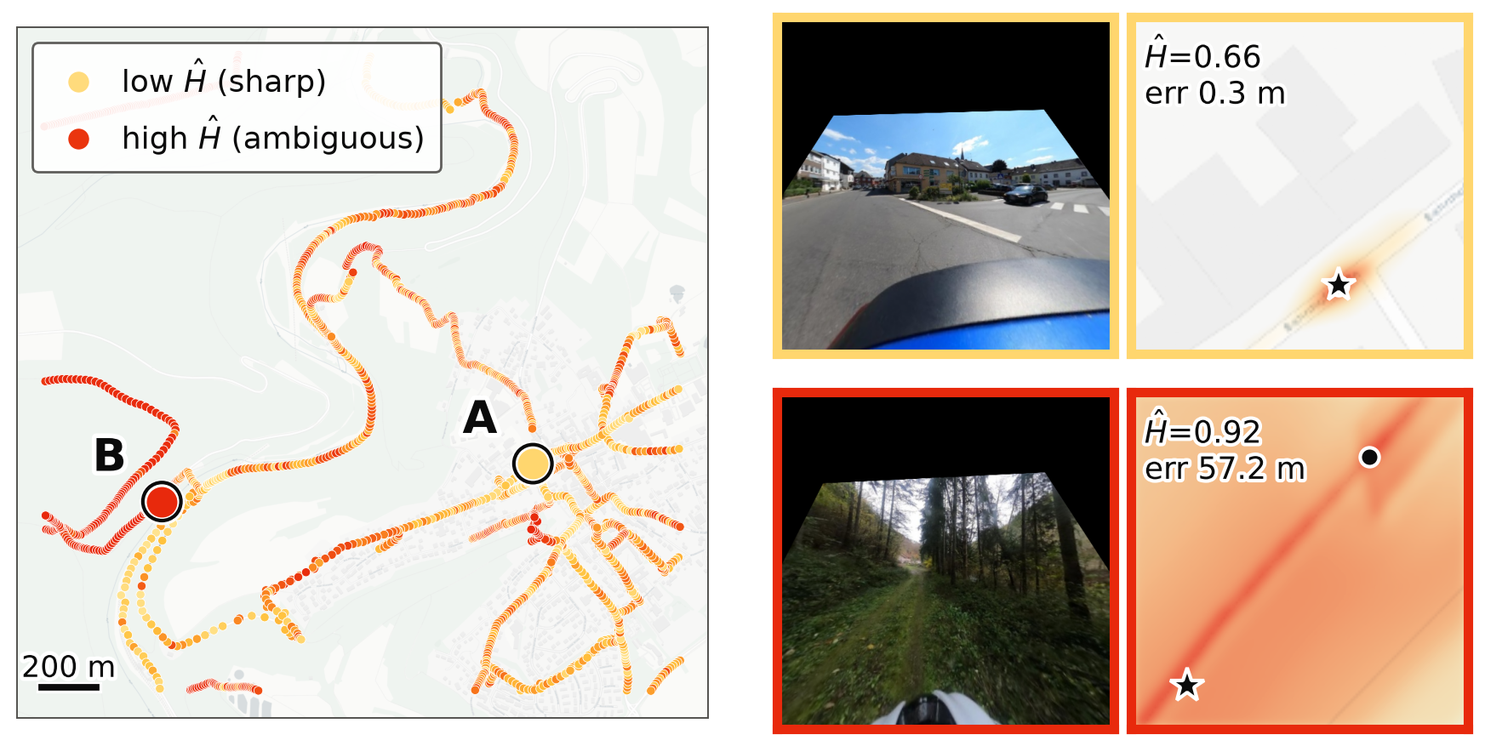}
\caption{Per-region entropy analysis on the Speicher region of CV-FSS, in the same format as Figure~\ref{fig:supp_etu_neuanspach}.}
\label{fig:supp_etu_speicher}
\end{figure}

\begin{figure}[t]
\centering
\includegraphics[width=\columnwidth]{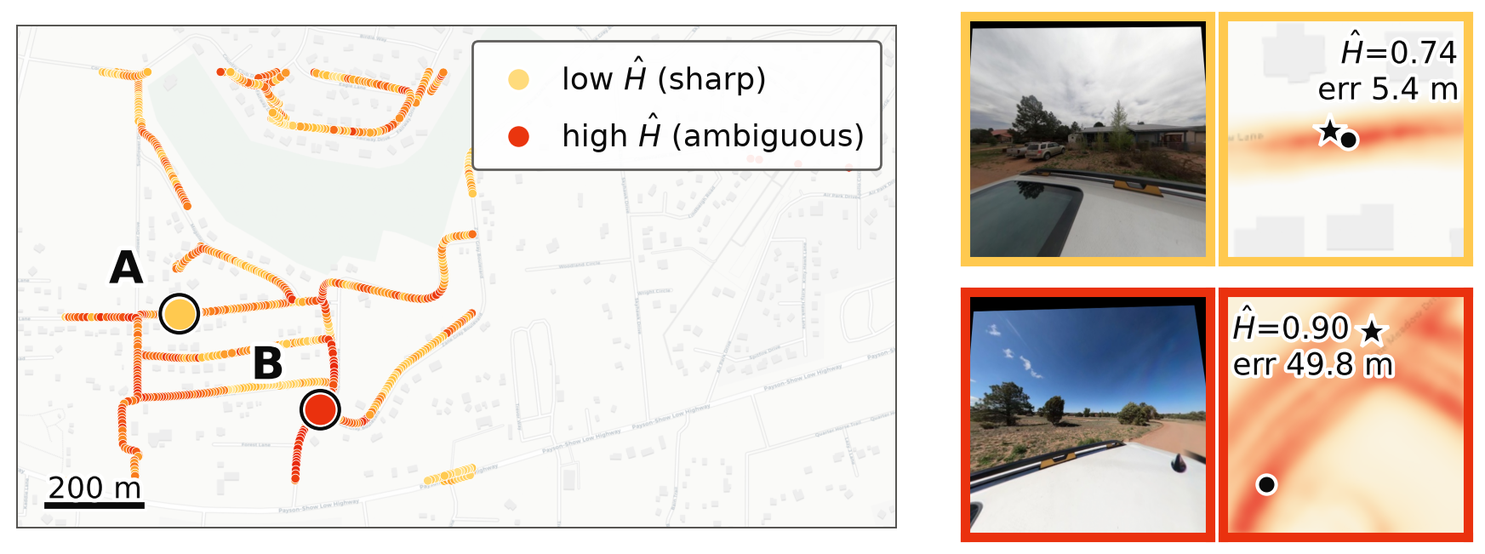}
\caption{Per-region entropy analysis on the Overgaard region of CV-FSS, in the same format as Figure~\ref{fig:supp_etu_neuanspach}.}
\label{fig:supp_etu_overgaard}
\end{figure}

\begin{figure*}[t]
\centering
\includegraphics[width=\textwidth]{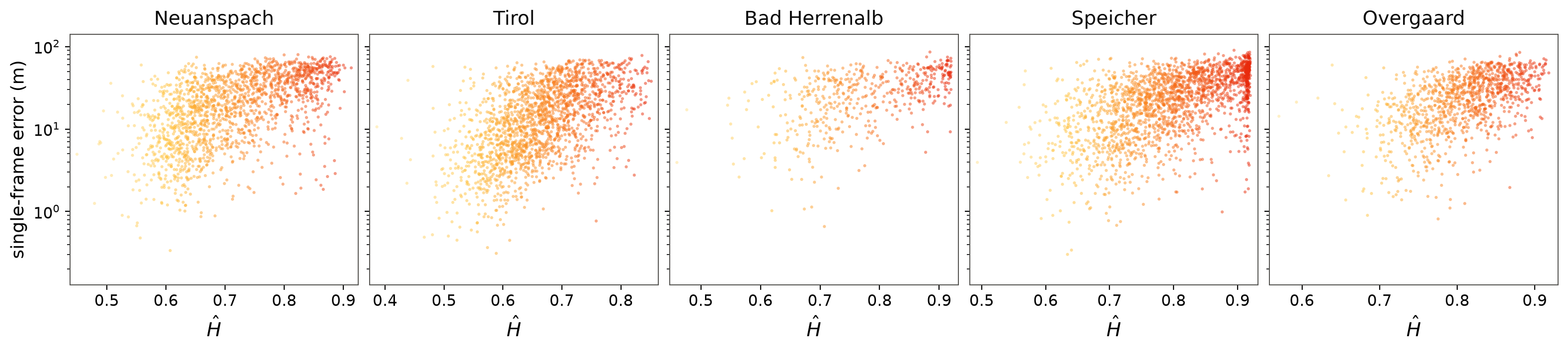}
\caption{Single-frame localization error against the normalized entropy $\hat{H}$ on the five regions of CV-FSS.
Each point is a test frame, colored by $\hat{H}$ from low (yellow) to high (red).
Across all regions, the error tends to grow with entropy, which supports using $\hat{H}$ as a reliability signal in ETU.}
\label{fig:supp_etu_scatter}
\end{figure*}

The main paper motivates Entropy-Tempered Uncertainty (ETU) on the Neuanspach region, showing that the normalized entropy $\hat{H}$ of a single-frame pose likelihood reflects its reliability.
Here we extend this analysis to all five regions of CV-FSS, shown in Figure~\ref{fig:supp_etu_neuanspach} to Figure~\ref{fig:supp_etu_overgaard}.
In each figure, the left map colors every frame of the test route by its normalized entropy $\hat{H}$, and the right panels contrast a distinctive location (A, low $\hat{H}$) against an ambiguous one (B, high $\hat{H}$).
The same pattern holds across all five regions.
Low-entropy frames produce peaked likelihoods whose argmax stays close to the ground truth, whereas high-entropy frames produce diffuse likelihoods whose argmax drifts far away.
Overall, $\hat{H}$ provides an informative per-frame confidence signal, where a lower value marks a frame whose argmax is more likely to fall near the true pose, even though a low entropy alone cannot guarantee a correct estimate.
Figure~\ref{fig:supp_etu_scatter} makes this trend quantitative, plotting the single-frame localization error against $\hat{H}$ for all test frames across the five regions.

\section{More Data Samples of CV-FSS}
\begin{figure*}[t]
\centering
\includegraphics[width=0.45\textwidth]{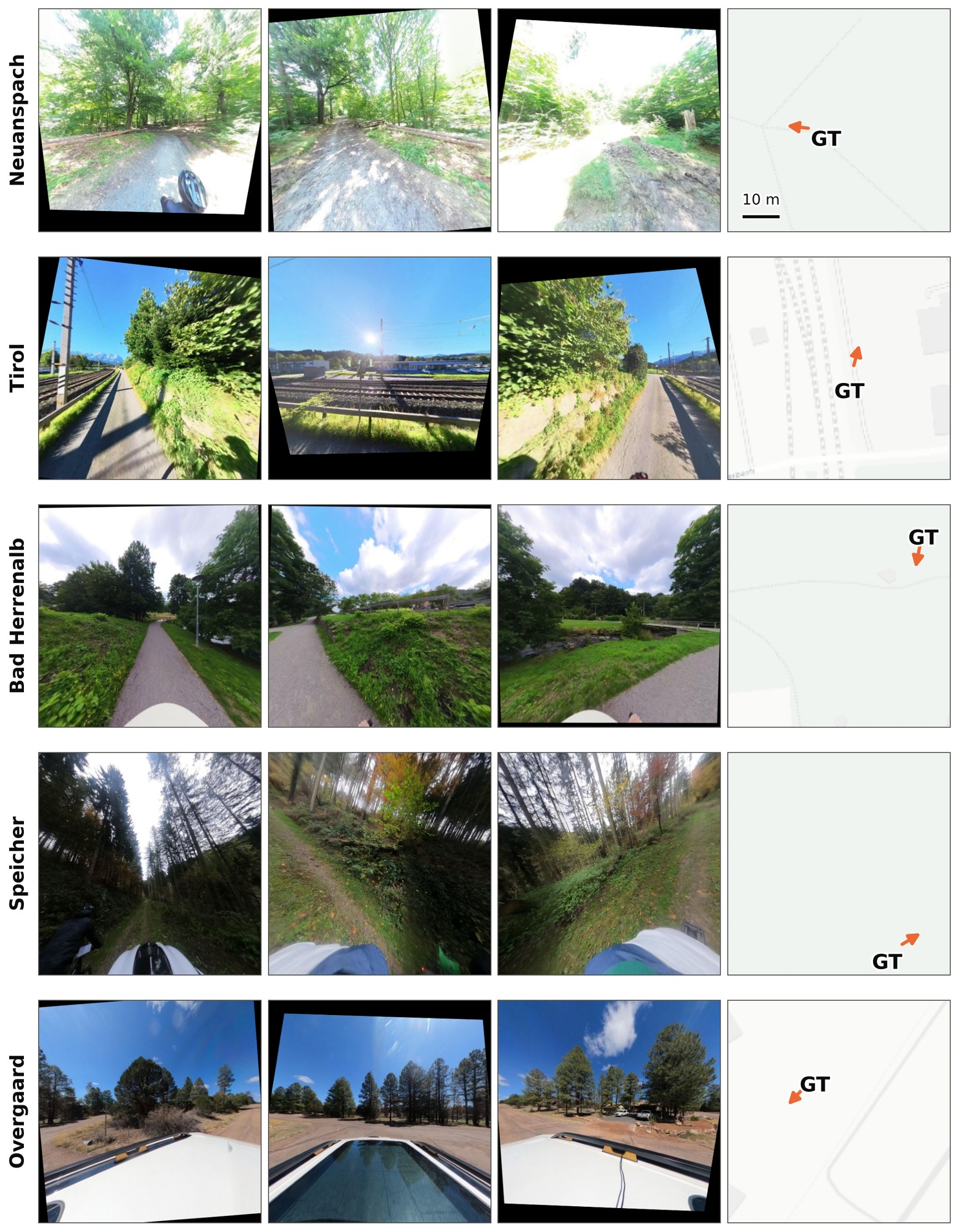}
\caption{More data samples from the five regions of CV-FSS.
From top to bottom, the rows correspond to the Neuanspach, Tirol, Bad Herrenalb, Speicher, and Overgaard regions.
The first three columns are the three perspective views split from one panorama, and the last column is the paired OSM tile, on which the orange arrow marks the ground-truth pose.}
\label{fig:supp_samples}
\end{figure*}

The main paper shows the structure of one CV-FSS sample.
Figure~\ref{fig:supp_samples} presents additional samples covering all five regions, one region per row.
Each panorama is split into three perspective views (the first three columns) that are rectified with the known gravity direction, and is paired with a local OpenStreetMap (OSM) tile (the last column) on which the ground-truth pose is marked.
Across regions, the surroundings are dominated by vegetation and road surface with few distinctive buildings or landmarks, so a single frame is often insufficient to determine the pose.

\section{Societal Impact}
SeqLoc improves self-localization when GNSS is unreliable, which can make autonomous vehicles and mobile robots safer in urban canyons, under signal jamming, or in remote areas with weak coverage.
Because it builds on OSM, a lightweight and freely available map, it reduces the reliance on satellite imagery or high-definition maps and extends metric localization to feature-sparse rural regions that are often overlooked.

Localization technology nonetheless carries risks.
Accurate positioning can be misused for surveillance or tracking, so any deployment should follow clear consent and data-protection rules.
Localization is also never perfect, and a wrong pose could mislead a safety-critical system, so SeqLoc should serve as one component within a multi-sensor stack rather than the sole source of pose.
Finally, its coverage depends on OSM, whose uneven quality across regions may introduce geographic bias.

\begin{table}[t]
\centering
\begin{tabular}{ll}
\toprule[1.5pt]
Setting & Value \\
\midrule
\multicolumn{2}{l}{\textit{Data and preprocessing}} \\
Map resolution & $2$ px/m \\
Map crop & $64$ m to $128\times128$ px \\
Input image & $512$, padded, pitch-rectified \\
Augmentation & rot90, flip, photometric \\
\midrule
\multicolumn{2}{l}{\textit{Architecture}} \\
Image encoder & ResNet-101 (ImageNet) \\
Map encoder & VGG-19 (from scratch) \\
BEV extent ($z$, $x$) & $32$ m, $55$ m \\
Latent / matching dim & $128$ / $8$ \\
Scale bins & $33$ \\
Parameters & $54.9$ M \\
\midrule
\multicolumn{2}{l}{\textit{Training}} \\
Optimizer & Adam (lr $10^{-5}$, wd $10^{-5}$) \\
LR schedule & ReduceLROnPlateau \\
Max epochs & $30$ \\
Effective batch & $36$ views ($12$ panoramas) \\
Rotation bins & $64$ \\
Loss & NLL over $H\times W\times R$ \\
Hardware & one $40$ GB A100 \\
\midrule
\multicolumn{2}{l}{\textit{Evaluation}} \\
Rotation bins & $256$ \\
Initial crop error & $32$ m \\
Augmentation & off \\
\bottomrule[1.5pt]
\end{tabular}
\caption{Training and evaluation configuration of the RHO backbone, trained from scratch on each scene.
The effective batch counts dataset items (views), and RHO packs three views per panorama, so its $36$ views correspond to $12$ panoramas.
The three test-time components of SeqLoc add no learned parameters and we use $\varepsilon = 0.01$ and $r = 3$.}
\label{tab:supp_config}
\end{table}

\begin{table}[t]
\centering
\setlength{\tabcolsep}{2.5pt}
\begin{tabular}{@{}cllcccccc@{}}
\toprule[1.5pt]
& & & \multicolumn{3}{c}{CV-FSS} & \multicolumn{3}{c}{CV-RHO} \\
\cmidrule(lr){4-6} \cmidrule(lr){7-9}
& Backbone & Mode & 1$\uparrow$ & 3$\uparrow$ & 5$\uparrow$ & 1$\uparrow$ & 3$\uparrow$ & 5$\uparrow$ \\
\midrule
\multirow{12}{*}{\rotatebox[origin=c]{90}{Pos.\ (m)}}
& \multirow{4}{*}{OrienterNet} & Single & 0.3 & 3.3 & 8.3 & 7.8 & 31.8 & 43.8 \\
& & Offline & 1.2 & 8.9 & 13.3 & 9.3 & 40.6 & 53.9 \\
& & Online & 1.9 & 10.5 & 17.1 & 10.2 & 42.7 & 54.4 \\
& & \textbf{SeqLoc} & \textbf{3.1} & \textbf{23.0} & \textbf{39.9} & \textbf{20.0} & \textbf{59.6} & \textbf{70.3} \\
\addlinespace[2pt]
& \multirow{4}{*}{OSMLoc} & Single & 0.4 & 3.6 & 8.8 & 8.5 & 33.6 & 46.0 \\
& & Offline & 2.0 & 11.8 & 18.3 & 10.7 & 43.9 & 55.4 \\
& & Online & 2.2 & 13.3 & 20.4 & 12.3 & 45.8 & 60.5 \\
& & \textbf{SeqLoc} & \textbf{7.1} & \textbf{32.3} & \textbf{49.8} & \textbf{21.5} & \textbf{62.4} & \textbf{73.5} \\
\addlinespace[2pt]
& \multirow{4}{*}{RHO} & Single & 0.6 & 4.8 & 10.6 & 9.0 & 35.3 & 47.9 \\
& & Offline & 2.7 & 14.4 & 20.5 & 21.1 & 49.2 & 63.9 \\
& & Online & 3.1 & 16.3 & 25.5 & 24.8 & 50.8 & 66.7 \\
& & \textbf{SeqLoc} & \textbf{14.5} & \textbf{55.9} & \textbf{75.0} & \textbf{37.6} & \textbf{86.8} & \textbf{94.2} \\
\midrule
\multirow{12}{*}{\rotatebox[origin=c]{90}{Ori.\ ($^\circ$)}}
& \multirow{4}{*}{OrienterNet} & Single & 3.6 & 11.4 & 18.0 & 15.0 & 39.3 & 53.6 \\
& & Offline & 7.5 & 20.7 & 32.3 & 20.9 & 44.6 & 56.9 \\
& & Online & 8.8 & 25.5 & 35.8 & 22.6 & 47.8 & 61.3 \\
& & \textbf{SeqLoc} & \textbf{24.5} & \textbf{49.4} & \textbf{60.1} & \textbf{39.8} & \textbf{69.0} & \textbf{77.2} \\
\addlinespace[2pt]
& \multirow{4}{*}{OSMLoc} & Single & 5.2 & 14.7 & 22.3 & 19.9 & 46.6 & 58.3 \\
& & Offline & 9.9 & 22.7 & 33.4 & 27.7 & 65.2 & 71.9 \\
& & Online & 10.8 & 29.2 & 37.6 & 28.2 & 70.4 & 77.5 \\
& & \textbf{SeqLoc} & \textbf{27.4} & \textbf{54.3} & \textbf{65.3} & \textbf{46.0} & \textbf{77.8} & \textbf{83.6} \\
\addlinespace[2pt]
& \multirow{4}{*}{RHO} & Single & 6.7 & 18.4 & 27.0 & 25.0 & 53.8 & 63.4 \\
& & Offline & 10.3 & 28.8 & 34.1 & 28.5 & 70.6 & 75.3 \\
& & Online & 12.2 & 32.1 & 38.8 & 30.1 & 76.7 & 79.3 \\
& & \textbf{SeqLoc} & \textbf{55.3} & \textbf{86.2} & \textbf{92.2} & \textbf{75.0} & \textbf{95.4} & \textbf{97.0} \\
\bottomrule[1.5pt]
\end{tabular}
\caption{Online versus offline test-time sequence aggregation for three backbones, averaged over the five regions of CV-FSS and the seven cities of CV-RHO.
We report position recall (\%) at 1/3/5\,m and orientation recall (\%) at 1/3/5$^\circ$.
Single is the single-frame backbone, offline sums all frames in one batch with future frames, online fuses only past frames recursively, and SeqLoc adds ETU, MGR, and PAS on top.}
\label{tab:supp_ttsa}
\end{table}

\FloatBarrier

\begin{table*}[t]
\centering
\begin{tabular}{lrrrrrrrrrr}
\toprule[1.5pt]
& & & \multicolumn{2}{c}{Mean error} & \multicolumn{3}{c}{Position recall (\%)} & \multicolumn{3}{c}{Orientation recall (\%)} \\
\cmidrule(lr){4-5}\cmidrule(lr){6-8}\cmidrule(lr){9-11}
Noise level & $\sigma_t$ (\%) & $\sigma_r$ ($^\circ$/m) & Pos.\ (m) & Ori.\ ($^\circ$) & R@$1$m & R@$3$m & R@$5$m & R@$1^\circ$ & R@$3^\circ$ & R@$5^\circ$ \\
\midrule
none (ideal) & $0$ & $0$ & $3.7$ & $2.6$ & $16.4$ & $60.7$ & $84.0$ & $63.7$ & $93.4$ & $96.8$ \\
good & $1$ & $0.005$ & $3.7$ & $2.6$ & $14.3$ & $59.8$ & $83.7$ & $61.8$ & $93.6$ & $96.8$ \\
medium & $2$ & $0.010$ & $3.8$ & $2.7$ & $13.7$ & $58.7$ & $82.9$ & $57.2$ & $93.6$ & $96.8$ \\
poor & $5$ & $0.020$ & $4.2$ & $2.8$ & $8.4$ & $49.1$ & $77.1$ & $45.6$ & $91.1$ & $96.9$ \\
very poor & $10$ & $0.050$ & $5.6$ & $4.5$ & $4.1$ & $34.5$ & $61.9$ & $27.4$ & $78.0$ & $92.4$ \\
\bottomrule[1.5pt]
\end{tabular}
\caption{Robustness to noisy odometry on the Neuanspach region of CV-FSS, with $\Delta_t$ perturbed by a zero-mean Gaussian noise at five levels.
Pos.\ and Ori.\ are the mean position and orientation errors of SeqLoc.
The recall columns report the percentage of frames localized within $1$, $3$, and $5$ m in position and within $1$, $3$, and $5^\circ$ in orientation.}
\label{tab:supp_odom}
\end{table*}

\end{document}